\documentclass{article}
\usepackage[letterpaper,top=2cm,bottom=2cm,left=3cm,right=3cm,marginparwidth=1.75cm]{geometry}
\usepackage{natbib}
\usepackage{parskip}

\usepackage{amsmath,amsfonts,bm}

\def\eqref#1{equation~\ref{#1}}

\def\1{\bm{1}}

\DeclareMathAlphabet{\mathsfit}{\encodingdefault}{\sfdefault}{m}{sl}
\SetMathAlphabet{\mathsfit}{bold}{\encodingdefault}{\sfdefault}{bx}{n}

\usepackage{xcolor}
\definecolor{linkpink}{HTML}{D33682}
\usepackage{url}

\usepackage{algorithm}
\usepackage{algorithmic}

\usepackage[utf8]{inputenc}
\usepackage{textcomp}
\usepackage{amssymb}
\usepackage{amsmath,amsfonts}
\usepackage{amsthm}
\usepackage{latexsym}
\usepackage{mathrsfs}
\usepackage{subfigure}
\usepackage{dsfont}
\usepackage{bbm}
\usepackage{multirow}
\usepackage{varioref}
\usepackage{booktabs}
\usepackage{bm}
\usepackage{graphicx}
\usepackage{amsopn}
\usepackage{wrapfig}
\usepackage{makecell}
\usepackage{hyperref}
\hypersetup{
    colorlinks=true,
    citecolor=blue,
    linkcolor=blue,
    urlcolor=linkpink
}

\DeclareRobustCommand{\modified}[1]{{#1}}

\newtheorem{remark}{Remark}

\graphicspath{{fig/}}

\newcounter{spb}
\def\RR{{\mathbb R}}
\def\PP{{\mathbb{P}}}
\def\cV{\mathcal{V}}
\def\1{\mathbbm{1}}

\newcommand{\Norm}[1]{\left\| #1 \right\|}

\title{Draft Calibration for Speculative Decoding: Accelerating LLM Serving by Matching Draft and Target Distributions}

\author{
\setcounter{footnote}{1}
Yuxing Liu$^{1}$\footnote{Work partly done during internship at Microsoft.} \quad Wei Zuo$^{2}$ \quad Walid Krichene$^{2}$ \quad Tong Zhang$^{1}$ \quad Li Zhang$^{2}$ \\ \\
{
$^{1}$UIUC}
\quad {$^{2}$Microsoft } 
}

\date{}

\begin{document}

\maketitle

\begin{abstract}
    Speculative decoding accelerates large language model (LLM) inference by employing a lightweight draft model to propose draft tokens and allowing the target model to verify the new tokens in parallel to improve inference-time compute utilization. 
    The speedup relies on the acceptance probability of draft tokens in target verification, which is directly related to the gap $\Norm{p-q}_1$ between the target distribution $p$ and the draft distribution $q$.
    In this paper, we study draft calibration of speculative decoding to improve the speedup of speculative decoding, where we employ post-hoc methods to calibrate the draft distribution to mitigate the distribution gap.
    We observe that though current community open-source draft models may learn the top candidate token correctly, they are generally underconfident compared to the target model, suggesting a consistent distribution mismatch.
    To mitigate the effect of this mismatch on the acceptance rate of the draft models, we try various post-hoc calibration methods, including standard temperature scaling and token-wise calibration. Our results suggest that a standard temperature scaling applied to the draft distribution improves the acceptance length by up to approximately 10\% with almost no additional cost across different tasks. \modified{Throughput analysis further verifies that this improvement in serving efficiency.} Analysis of more general calibration methods reveals further potential room for improvement.
\end{abstract}

\section{Introduction}
Current state-of-the-art large language models (LLMs) generate text in an autoregressive way, which runs one new forward pass to generate one next token recursively. This nature of LLM serving results in an insufficient use of parallelism in modern computational resources like GPUs. 
Speculative decoding~\citep{leviathan2023fast,chen2023accelerating} is considered a valid solution to mitigate this fundamental serving inefficiency: at each inference time, a lightweight draft model first proposes a draft block that contains several draft tokens in a row, and the target model (the serving model) verifies all the draft tokens through one forward pass. 
Thanks to the rejection sampling design at the verification stage, speculative decoding perfectly preserves the target model distribution while accelerating inference by lowering the number of target model forward passes to improve compute utilization and ultimately serving efficiency.

The speedup achieved by speculative decoding fundamentally relies on the acceptance rate of each draft token, i.e., the probability that a draft token can be accepted by the target model verification, which has a closed-form representation $1-\frac{1}{2}\Norm{p-q}_1$ on target distribution~$p$ and draft distribution~$q$~\citep{leviathan2023fast,chen2023accelerating}. 
Therefore, a clear target to improve speculative decoding is that we not only desire the draft model to learn top candidates for each token well, but also to minimize this $\ell_1$ distribution gap. 
This formula motivates a connection to confidence calibration~\citep{guo2017calibration,wang2023calibration}, where predicted probabilities should match empirical correctness frequencies rather than merely identify the top candidate class. In speculative decoding, the formulation can be more straightforward: 
since the goal is to improve the acceptance rate, we consider the target distribution as the ground truth, and calibrate the draft distribution to mitigate this $\ell_1$ distribution gap and thus improve the acceptance rate. 
From this perspective, we may improve speculative decoding not only in training time through algorithm and architecture modifications~\citep{cai2024medusa,li2024eagle,li2024eagle2,li2025eagle3,chen2026dflash,cheng2026dspark}, but also after training by calibrating the draft distribution better, which we study systematically across draft architectures.

In this paper, we study this draft calibration idea of speculative decoding, mainly on post-hoc methods adjusting the draft distribution $q$ to mitigate the $\ell_1$ distribution gap at inference time when using chain draft generation. 
In existing literature and community speculative decoding serving frameworks, one usually shares the inference settings as the target model without any modification when chain generating draft tokens.
However, as we observe across community open-source draft models on different tasks, this choice may not be optimal because of the existence of a systematic mismatch between the draft and target models under the original setup. 
More specifically, the draft models tend to be underconfident compared to the target model when using the same generation settings, even though they have learned the top candidates correctly, revealing a consistent distribution gap. 

To mitigate this gap and improve the speculative decoding speedup, we consider multiple calibration methods applied to the draft distribution, and show that a standard temperature scaling~\citep{guo2017calibration} for the draft distribution can significantly mitigate the gap and improve the overall accepted length by up to around 10\% with almost no additional cost. We also further study Dirichlet scaling~\citep{kull2019beyond}, a generalization of temperature scaling, and token-adaptive calibration methods, which apply more fine-grained adjustments to the draft distribution. Analysis of the optimal choices for these methods reveals further potential room of improvement with draft calibration, but currently the tried post-hoc methods still cannot capture it well.

Our key contributions include:
\begin{itemize}
    \item We present the underconfident tendency of draft models compared to target models, forming a consistent distribution mismatch that affects acceptance rate.
    \item We study the problem from the draft calibration perspective and apply different calibration methods to mitigate the systematic distribution gap, showing that standard temperature scaling can already significantly improve the acceptance rate across different settings.
    \item We further examine more general calibration methods and reveal that there is room for improvements with token-adaptive calibration methods through a trajectory replay analysis framework, though the current pure post-hoc methods may not be able to capture it.
\end{itemize}

\section{Backgrounds and Understanding of Draft Calibration}\label{sec:backgrounds}
\textbf{Speculative decoding.}
To accelerate the inference of a large model, referred to as the target model, speculative decoding~\citep{leviathan2023fast,chen2023accelerating} incorporates a lightweight draft model and follows a draft-then-verify pipeline. At inference time, given a prefix sequence $x_1,\dots,x_t$, we let the draft model generate $K$ draft tokens in a row, then feed all the draft tokens to the target model to run a forward pass to verify them. The verification process goes one by one, i.e., if one draft token is not accepted by the verification, we abandon all of the subsequent draft tokens. For each draft token $x_t$ at position $t$ with draft distribution $q$ and target distribution $p$ on vocabulary $\cV$, we do a rejection sampling process: we accept the token if $p(x_t) \ge q(x_t)$, otherwise we reject it with probability $1 - \frac{p(x_t)}{q(x_t)}$ and resample a new token from an adjusted distribution $p'(x) = \mathrm{normalize}(\max(0,p(x) - q(x)))$. We can compute the expected acceptance rate of each draft token from the sampling strategy:
\begin{align}\label{eq:acceptance_probability}
    \PP(x_t \text{ is accepted}) = \sum_{x \in \cV} \min(p(x),q(x)) = 1 - \frac{1}{2} \Norm{p-q}_1 .
\end{align}
Clearly, mitigating this $\ell_1$ gap between draft and target models directly improves acceptance rate and the effectiveness of speculative decoding.

\textbf{Confidence calibration and draft calibration.} 
Confidence calibration has been an important field for deep learning and trustworthy AI~\citep{guo2017calibration}. In a classification problem of feature label pairs $(X,y)$, calibration measures how much a model's predicted probability matches the true correctness likelihood. More formally, a classifier model that maps the feature $X$ to a probability simplex $q$ is perfectly calibrated if $\PP(y = y_i | q(y_i) = q_i) = q_i$ for class $i$. Post-hoc methods fit a calibration map on held-out data while keeping the predictor fixed, ranging from histogram binning and isotonic regression to temperature and Dirichlet scaling~\citep{guo2017calibration,kull2019beyond}. 

Given the goal of improving acceptance of draft tokens for speculative decoding, we can also consider it from the calibration perspective, where we try to mitigate the distribution gap between the draft distribution and target distribution, and we term it \textbf{draft calibration} in this paper. Compared to the confidence calibration problem, the idea is more straightforward: we have the ground truth target distribution after verification for each draft token, and can simply adopt acceptance performance as the evaluation metric for draft calibration.
In this paper, we will focus on post-hoc methods to help calibrate the draft distribution better, which preserves the lossless property of speculative decoding, keeping the target distribution unchanged as long as the calibrated distribution $q$ is employed consistently in drafting and rejection sampling stages~\citep{leviathan2023fast,chen2023accelerating}. 

\textbf{Relation to draft tree methods.}
We focus on chain drafting in this paper, motivated by large-scale serving, where we handle many concurrent requests that compete for verification compute. Draft trees increase candidate coverage by proposing multiple continuations at each draft position~\citep{li2024eagle2,li2026dartree}, but their additional branches also introduce more verification costs. This tradeoff becomes more restrictive as batch size grows: \citet{liu2026speculative}, for example, find that a 21-token tree becomes slower than autoregressive decoding at batch size 64 across their evaluated workloads, while the three-token chain retains a speedup. Likewise, the concurrent-serving study in \citet{li2026dartree} also reports largely reduced tree node budgets to keep the effectiveness of the draft tree with a batch size of 32. The evidence suggests that the benefits of draft trees may be limited in large-scale serving, where the batch size can be up to hundreds.
We therefore keep our focus on how much acceptance can be improved by calibrating the draft distribution. 

\section{Related Work}

\textbf{Speculative decoding.}
Early blockwise and speculative methods amortize verification over multiple candidate tokens~\citep{stern2018blockwise,xia2023speculative}, while speculative sampling pairs a lightweight drafter with parallel target verification to preserve the target distribution~\citep{leviathan2023fast,chen2023accelerating}. Subsequent work improves draft efficiency and alignment through alternative backbones and distillation~\citep{choi2025mamba,zhou2024distillspec,zhang2025hass}, and expands candidate coverage through tree-based verification~\citep{miao2024specinfer,li2024eagle2,li2026dartree}. Medusa, EAGLE, and related draft-head designs introduce additional decoding heads or a lightweight draft layer conditioned on target features instead of standalone draft models~\citep{cai2024medusa,li2024eagle,ankner2024hydra}, and the latter state-of-the-art methods follow this design, including self-speculation methods~\citep{zhang2024draftverify,liu2024kangaroo} and EAGLE-3~\citep{li2025eagle3}. ParallelSpec, P-EAGLE, DFlash, and related parallel drafters reduce sequential drafting overhead by predicting multiple positions in parallel, including through block-diffusion drafting~\citep{xiao2024parallelspec,hui2026peagle,an2026pard,chen2026dflash}. Built upon the parallelism idea, DSpark augments a parallel backbone with lightweight autoregressive correction and confidence-based verification scheduling~\citep{cheng2026dspark}. 

\textbf{Confidence calibration.}
Confidence calibration aligns predicted confidence with empirical correctness and has been studied through post-hoc maps and calibration-aware training~\citep{guo2017calibration,kumar2018trainable,mukhoti2020calibrating,wang2023calibration}. Post-hoc approaches include binning, parametric scaling, and their compositions~\citep{naeini2015obtaining,kull2017beta,kull2019beyond,kumar2019verified,zhang2020mix,ding2021local}. More recent studies also examine token-adaptive temperature scaling for language models~\citep{desai2020calibration,xie2024calibrating}. 

\textbf{Post-hoc draft calibration in speculative decoding.}
Draft calibration is largely unexplored in the literature, and prior works with temperature and distribution adjustment
address several distinct objectives~\citep{wang2025subspec,ouyang2024temperature,zhang2026flexdraft,berdoz2026steering,zhou2026calibrated}, with the following works as examples. SubSpec sharpens draft probabilities to improve cumulative-probability ranking in deep draft trees for offloaded inference with a greedy target, but
does not systematically study the draft calibration~\citep{wang2025subspec}. \citet{ouyang2024temperature} study the interaction between the training data of the draft model and decoding temperatures, but does not consider differentiating the temperature of the draft model. Other work calibrates confidence for control decisions: SpecBound temperature-scales shallow-layer outputs for early exit, while DSpark calibrates predicted prefix acceptance for verification scheduling~\citep{wen2026specbound,cheng2026dspark}. Differently, we focus on systematically studying the distribution gap between draft and target models and methods to mitigate the gap.

\section{Underconfident Draft Models}\label{sec:underconfident_draft_models}

We start from examining the current popular open-source draft models' distribution gap compared to the corresponding target models, which forms the basis of applying post-hoc draft calibration methods. While \citet{guo2017calibration} discovers that modern neural networks tend to be overconfident in the choice they make, we find that draft models are generally underconfident compared to target models, admitting a consistently lower probability for top candidates. In other words, the draft distribution is generally flatter.

\subsection{Experiment Settings \& Metrics}\label{sec:experiment_settings}
Unless otherwise specified, we employ the following experiment settings for speculative decoding evaluation throughout the paper.
We evaluate nine target--draft pairs using community open-source EAGLE-3~\citep{li2025eagle3}, DFlash~\citep{chen2026dflash}, and DSpark~\citep{cheng2026dspark} checkpoints for our evaluation. Target Models include GPT-OSS-120B, Gemma-4-31B-IT, and the Qwen series. Table~\ref{tab:model_settings} lists the exact pairings, checkpoint links, and chain lengths, which we generally follow the recommended settings. We use eight benchmarks spanning math (GSM8K, MATH-500, AIME 2025), code (HumanEval, MBPP, LiveCodeBench), and chat (MT-Bench, Arena-Hard v2), with sources and citations listed in Table~\ref{tab:dataset_settings}. We use low-reasoning/no-thinking chat template in the experiments unless otherwise specified.
The experiments are done on  SpecForge~\citep{li2026specforge}/SGLang~\citep{zheng2024sglang} stack with independent draft-temperature control and use chain proposals and rejection sampling with the calibrated draft probabilities. The temperature for the target model is fixed at $T_t=1$, and generation is capped at 2,048 tokens. Each model pair keeps its chain length fixed across calibration methods. Please refer to more evaluation details in Appendix~\ref{appendix:experiment_settings}.

We use the accepted length as our evaluation metric: output tokens committed per verification round, including the bonus token. We aggregate rounds within each task and report an equal-weight mean across tasks. Distribution diagnostics average over eligible draft positions, where all preceding proposals in the block were accepted.

\subsection{Draft Models can learn top candidates well, but are underconfident}
We use the same inference setup for draft and target models as default in this section.

\begin{figure}[H]
    \centering
    \vspace{-0.25cm}
    \includegraphics[width=1.0\linewidth]{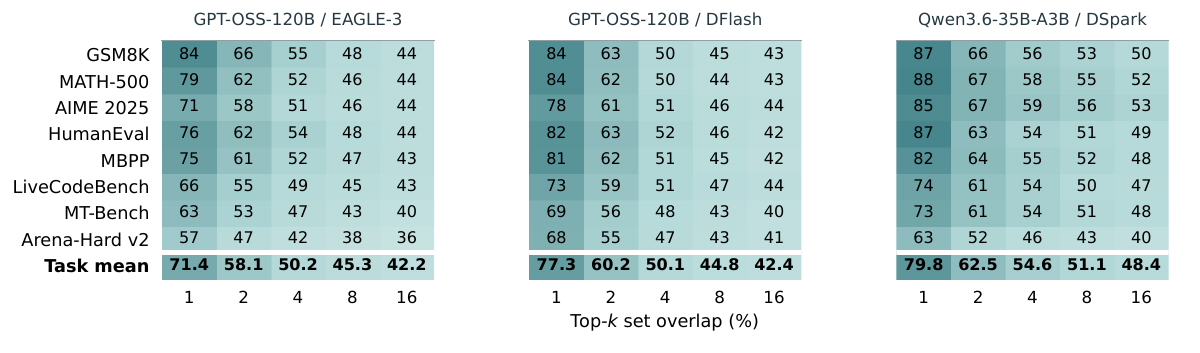}
    \vspace{-0.9cm}
    \caption{The overlap percentage of top-k candidates in target and draft distributions on three different draft models, where $k=1,2,4,8,16$.}
    \vspace{-0.55cm}
    \label{fig:topk_set_overlap}
\end{figure}

\textbf{Generally consistent top candidates.} We start from examining how consistent the top candidates learned by the draft model are with the target model. In Figure~\ref{fig:topk_set_overlap}, we show the overlap percentage between the top-k candidates of draft and target distributions regardless of order, i.e., how many of the top-k candidates are the same between target and draft models. We can observe that the draft model can learn around 70\% or more for the top-1 token, showing its capability for catching the top candidate in the target distribution. The correctness tends to decrease, but it still remains more than 40\% up to the top 16 candidates.

\begin{figure}[t]
    \centering
    \includegraphics[width=1.0\linewidth]{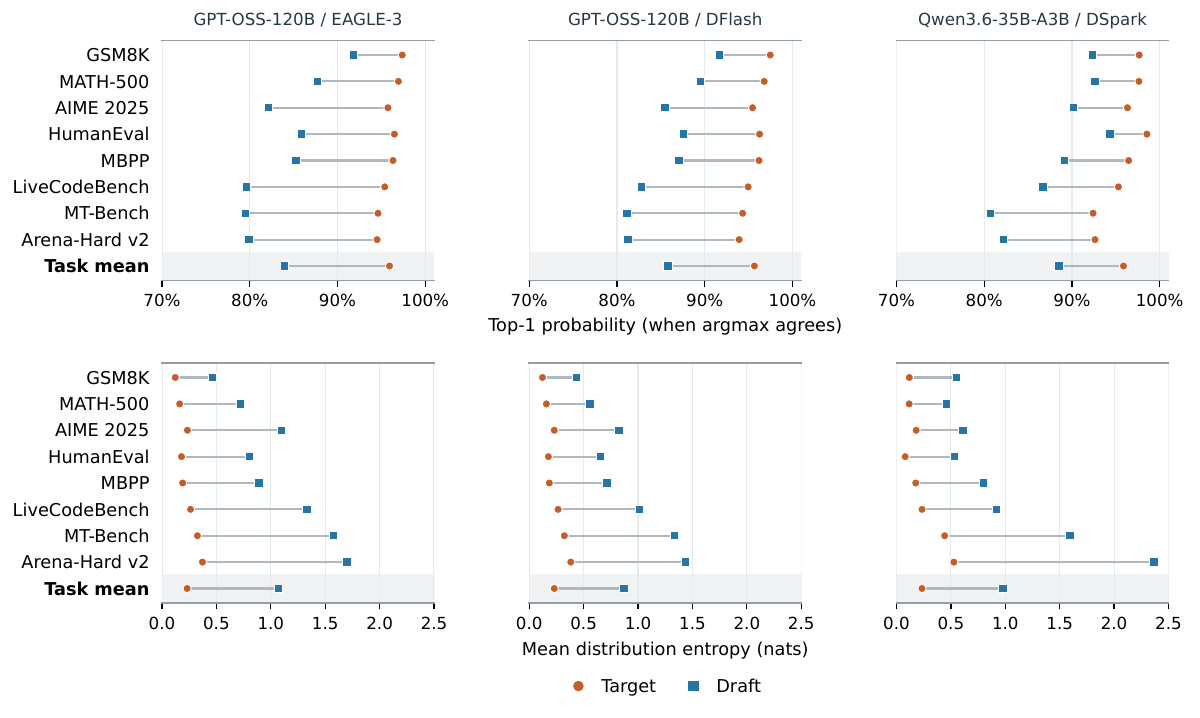}
    \vspace{-0.8cm}
    \caption{Top-1 candidate probability when the top candidate agrees (first row) and entropy (second row) of draft and target distributions on three different draft models. }
    \label{fig:matched_top1_prob_ent}
    \vspace{-0.3cm}
\end{figure}

\textbf{Underconfident draft model.} 
While the top candidates are generally learned correctly, we observe a consistent gap between the draft and target distributions. Figure~\ref{fig:matched_top1_prob_ent} shows the probability draft and target distributions give the top-1 candidate when they agree, and the entropy of the distributions. 
Across different tasks and state-of-the-art speculative decoding algorithms, there exists a consistent gap between the top-1 probability of draft and target distributions, even if the draft model predicts it correctly -- the draft model is underconfident about its top candidate choice compared to the target distribution. The consistently higher entropy also provides further evidence that the draft distribution is generally flatter than the target distribution. To conclude, the draft models are overall underconfident compared to the target model.

\begin{remark}[Why this distribution gap exists]\label{rem:distribution_gap}
    This consistent distribution gap is actually surprising since the draft distribution should be trained to match the target distribution. We hypothesize that this may be related to the limited capacity of draft models, making it hard to learn a sharp distribution as the target model. \modified{The settings of the training recipe could also be a potential reason for this. As discussed in Section~\ref{sec:temperature_scaling}, DeepSeek-trained draft models have a smaller draft-target distribution gap across different speculative decoding algorithms.} For example, the KL loss employed by the standard draft model training pipeline may also contribute to this, since it biases from the $\ell_1$ distribution gap~\citep{samarin2026lk}. 
\end{remark}

\section{Draft Calibration Methods}

Recall that the acceptance probability of a draft token can be computed by $A = 1 - \frac{1}{2} \Norm{p-q}_1$ that is directly related to the draft and target distribution gap.
Observing the underconfident and flat draft distribution, it is natural to consider post-hoc calibration to mitigate its gap with the target distribution to improve the acceptance performance. In this section, we examine from standard general calibration methods to token-wise adaptive calibration, and show that the simple and popular temperature scaling already gives a significant improvement in accepted length.

\subsection{Temperature Scaling}\label{sec:temperature_scaling}

Temperature scaling~\citep{guo2017calibration} has become a standard default calibration and LLM inference technique. Given logits $z \in \RR^V$ and temperature $T$, the corresponding probability distribution is:
\begin{align*}
    p_{\mathrm{cal}} = \mathrm{softmax}\left( \frac{z}{T} \right), \text{ i.e., } [p_T]_i = \frac{\mathrm{e}^{z_i/T}}{\sum_{j=1}^V \mathrm{e}^{z_j/T}} \text{ for any } i \in [V].
\end{align*}
From this formula, it is clear that lowering the temperature leads to a sharper distribution while keeping the rank of candidates the same. 
For speculative decoding, we actually have two temperature parameters: $T_t$ for the target model that affects the quality of the generated text, which is considered fixed in the speculative decoding problem, and $T_d$ for the draft model that only affects the draft distribution $q$, treated as an independent calibration parameter allowed to be different from $T_t$.

As presented in Section~\ref{sec:underconfident_draft_models}, the draft models are consistently underconfident with this default choice $T_d = T_t$ when $T_t = 1$. Since lowering $T_d$ leads to higher top candidate probability and sharper draft temperature, our first trial for post-hoc calibration is tuning this $T_d$ as a hyperparameter to check whether it helps mitigate the distribution gap, where we employ accepted length as the evaluation metric because of the direct relation between acceptance rate and distribution gap. The following experiments are conducted under the specified settings in Section~\ref{sec:experiment_settings}.

\begin{figure}[t]
    \centering
    \includegraphics[width=1.0\linewidth]{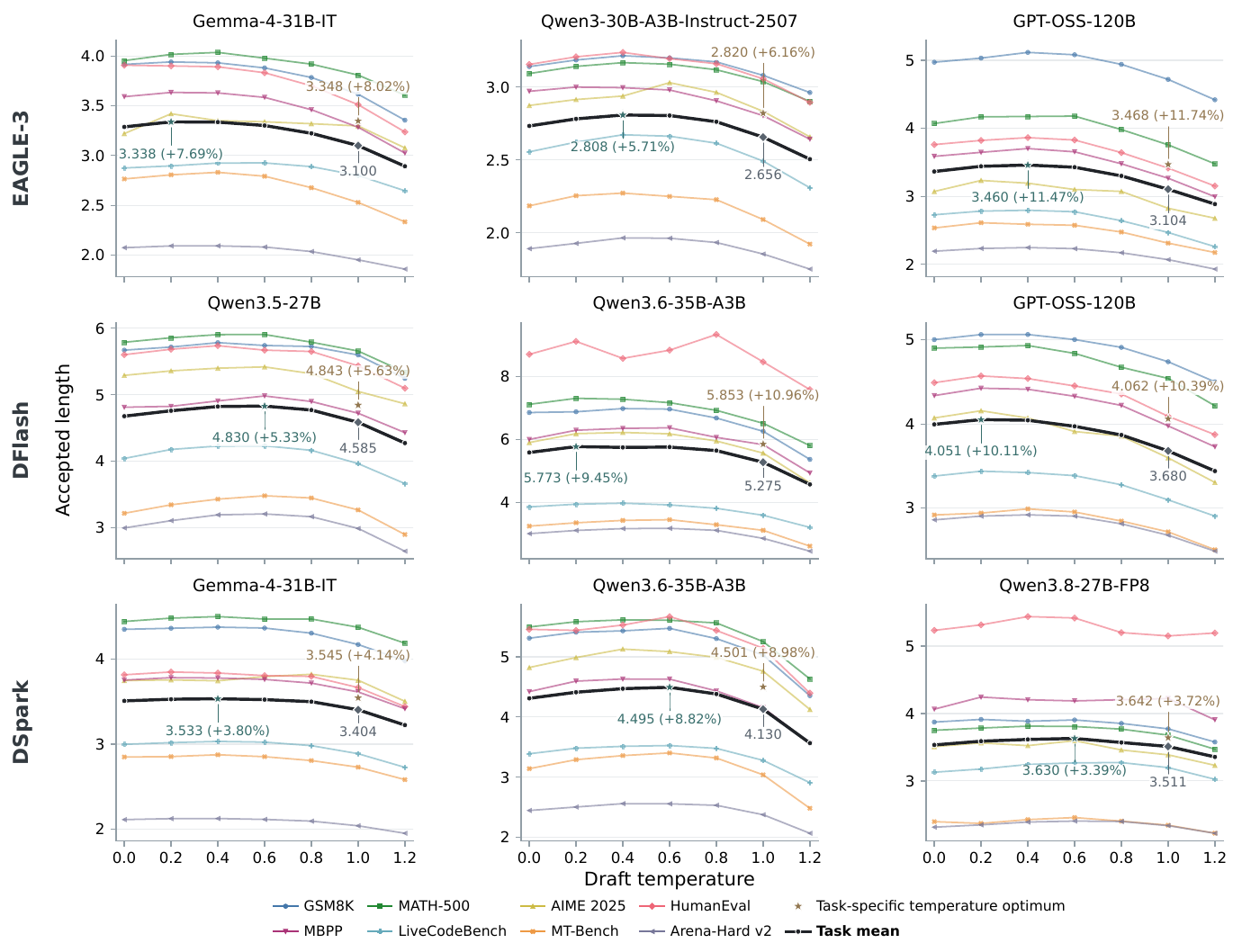}
    \vspace{-0.8cm}
    \caption{The accepted length across different models and speculative decoding algorithms when different draft temperatures are employed. The black line represents the average accepted length when the draft temperature is employed across all the tasks, and the bronze star means the optimal accepted length when we select the optimal temperature for each task. We specifically highlight their values and the baseline $T_d = 1$ for reference. 
    }
    \label{fig:temperature_scaling_accepted_length}
    \vspace{-0.3cm}
\end{figure}

\textbf{Temperature scaling shows improvements across models and tasks.}
We apply temperature scaling to the draft model to adjust its distribution with a grid of $T_d \in \{ 0,0.2,0.4,0.6,0.8,1,1.2 \}$ while keeping the temperature for the target model $T_t = 1$. Following the settings in Section~\ref{sec:experiment_settings}, we are using no/low thinking mode for the models.
We summarize the key observations on the results presented in Figure~\ref{fig:temperature_scaling_accepted_length}.
\begin{itemize}
    \vspace{-0.3cm}
    \item Across all the considered settings, accepted length shows a clear peak when $T_d \in [0,0.8]$. This is consistent with the observation that the draft models are generally underconfident, since $T_d < 1$ sharpens the draft distribution and achieves better performance. 
    \vspace{-0.2cm}
    \item When looking at the curves in Figure~\ref{fig:temperature_scaling_accepted_length}, we can find that: 1) across all the settings, accepted length is approximately unimodal with respect to the draft temperature, making it easier to tune; 2) the curves are quite flat when $T_d$ is near the optimal choice, implying that the current draft temperature grid should be sufficient for evaluating temperature scaling; 3) the trend of the curves for different tasks are relatively consistent for the same draft model, implying consistency in the distribution gap observed in Section~\ref{sec:underconfident_draft_models} across domains, though the curves for different models may be more different.
    \vspace{-0.2cm}
    \item We can observe that temperature scaling improves the average accepted length by approximately 5-10\%, e.g., from $5.275$ to $5.773$ for Qwen3.6-35B-A3B with DFlash, when employing the same draft temperature across {tasks for each model pair}, the improvement consistently exists across different tasks. For DSpark, there is also a 3\%-9\% gain in average accepted length, with a generally larger $T_d$. Changing the draft temperature adds no model computation, so accepted-length gains imply proportional decoding-throughput gains when the time per draft--verify round remains unchanged.
    \item We would like to further emphasize that temperature scaling is powerful without the need for careful tuning, since the curves are generally flat across a large temperature range. When draft temperature is set to $T_d \in [0.4,0.6]$, we can observe consistent gains compared to the standard choice $T_d = 1$ across all the models and tasks, which is generally near the best performance with temperature scaling.
    Task-specific draft temperature selection also gives only marginal improvement, suggesting that one global draft temperature already benefits multiple domains of use.
\end{itemize}
\vspace{-0.2cm}

\textbf{\modified{Throughput analysis provides direct evidence for serving efficiency.}}
\modified{We also report the throughput results for temperature scaling in Table~\ref{tab:serving_results}. Consistent with the fact that temperature scaling introduces almost no additional costs, the throughput analysis also implies significant gain across different request concurrency, providing direct evidence of serving efficiency improvements of draft calibration. Appendices~\ref{appendix:serving_protocol} and~\ref{appendix:serving_results} give the settings, explain the gain differences, and report latency and per-task results. Note that the throughput improvement does not exactly equal the accepted length improvement, which can be related to the usage of continuous batching.}

\begin{table}[t]
\centering
\small
\setlength{\tabcolsep}{4pt}
\caption{\modified{Online serving results on the held-out test split across all eight tasks, averaged over three seeds. Both metrics average tasks equally within each seed. Parentheses show mean per-seed gains over $T_d=1$; $C$ is total concurrency across two single-GPU replicas. Please refer to Table~\ref{tab:serving_uncertainty} for the full mean $\pm$ standard error results.}}
\label{tab:serving_results}
\vspace{0.1cm}
\modified{
\begin{tabular}{@{}lrrrrr@{}}
\toprule
& & \multicolumn{2}{c}{Accepted Length} & \multicolumn{2}{c}{Throughput (output tok/s)} \\
\cmidrule(lr){3-4}\cmidrule(l){5-6}
Target / drafter & $C$ & Original & Calibrated (gain) & Original & Calibrated (gain) \\
\midrule
\multirow{3}{*}{\makecell[l]{GPT-OSS-120B\\EAGLE-3}} & 16 & 3.102 & 3.432 (10.64\%) & 760.9 & 836.2 (9.92\%) \\
 & 64 & 3.104 & 3.436 (10.71\%) & 1582.6 & 1756.3 (11.01\%) \\
 & 128 & 3.103 & 3.441 (10.89\%) & 2141.6 & 2249.7 (5.13\%) \\
\midrule
\multirow{3}{*}{\makecell[l]{GPT-OSS-120B\\DFlash}} & 16 & 3.657 & 3.996 (9.27\%) & 1449.5 & 1592.5 (9.87\%) \\
 & 64 & 3.656 & 3.994 (9.25\%) & 2765.9 & 3034.3 (9.68\%) \\
 & 128 & 3.660 & 3.998 (9.26\%) & 3548.1 & 3866.0 (8.98\%) \\
\midrule
\multirow{3}{*}{\makecell[l]{Qwen3.6-35B-A3B\\DSpark}} & 16 & 4.049 & 4.434 (9.51\%) & 1174.9 & 1313.9 (12.04\%) \\
 & 64 & 4.066 & 4.425 (8.82\%) & 1858.6 & 2265.8 (22.08\%) \\
 & 128 & 4.070 & 4.430 (8.84\%) & 1782.9 & 2130.7 (19.53\%) \\
\bottomrule
\end{tabular}}
\vspace{-0.3cm}
\end{table}

\begin{figure}
    \centering
    \includegraphics[width=1.0\linewidth]{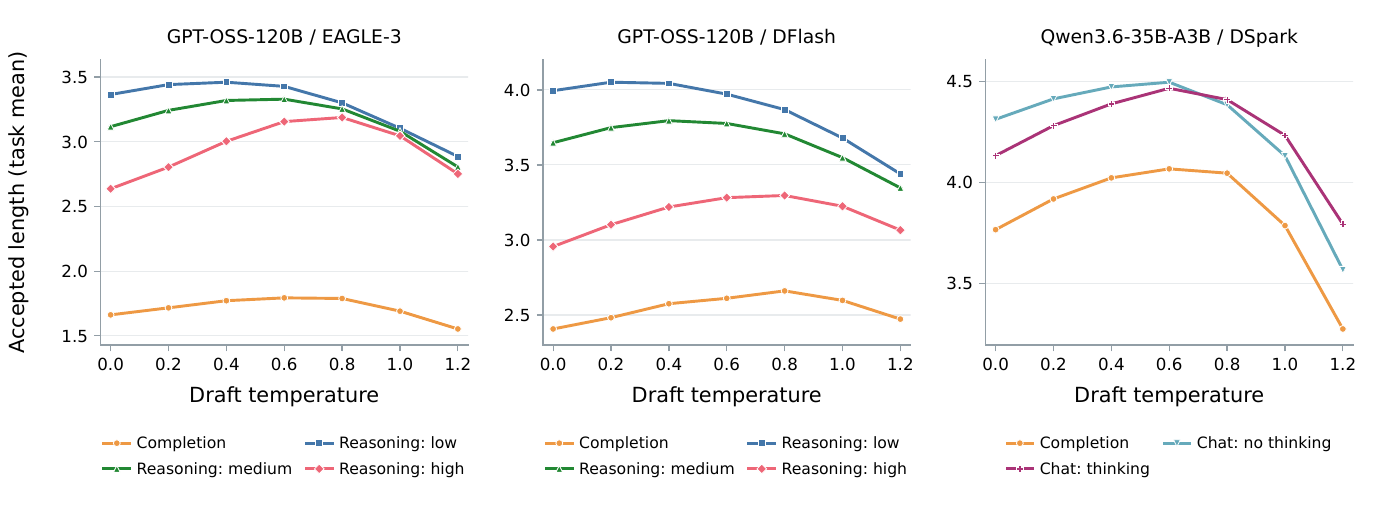}
    \vspace{-0.8cm}
    \caption{The accepted length averaged across the eight benchmarks with respect to draft temperature, with different chat templates.}
    \label{fig:temperature_scaling_eval_settings}
    \vspace{-0.3cm}
\end{figure}

\textbf{Reasoning effort (template) significantly affects speculative decoding.}
We further examine the performance of the draft models and the effect of temperature scaling when different evaluation templates are considered. Specifically, we evaluate the accepted length when serving with different reasoning effort as well as a completion mode, where we remove the chat template and let the model complete the context after the prompt. The results are presented in Figure~\ref{fig:temperature_scaling_eval_settings}. We can observe that the acceptance performance varies a lot across different reasoning efforts or evaluation templates, as also noticed in \citet{eldenk2026attention}.
Consistent with the low reasoning case in Figure~\ref{fig:temperature_scaling_accepted_length}, the optimal draft temperature still biases from $T_d = 1$ across all the evaluation settings.
The optimal draft temperature tends to increase mildly when the reasoning effort increases, which can be relevant to that the target distribution is itself generally flatter with a higher reasoning effort. Overall, temperature scaling benefits all the setups and generally gives more improvement in lower reasoning settings.

\begin{wraptable}{r}{0.48\textwidth}
    \centering
    \small
    \setlength{\tabcolsep}{3pt}
    \vspace{-0.6cm}
    \caption{\modified{DeepSeek-trained Qwen3-14B drafters. AL denotes accepted length; gains use the best shared $T_d$ over $T_d=1$. Top-1 gap is target minus draft probability at $T_d=T_t=1$ on argmax agreement (percentage points). All metrics average eight tasks equally.}}
    \vspace{0.1cm}
    \label{tab:deepseek_summary}
    \modified{\begin{tabular}{@{}lccc@{}}
    \toprule
    Drafter & Best $T_d$ & AL gain & Top-1 gap \\
    \midrule
    EAGLE-3 & 0.6 & 2.05\% & 5.65 \\
    DFlash & 0.8 & 0.10\% & 0.93 \\
    DSpark & 1.0 & 0\% & 0.18 \\
    \bottomrule
    \end{tabular}}
    \vspace{-0.2cm}
\end{wraptable}
\textbf{\modified{DeepSeek-trained draft models are better calibrated.}}
\leavevmode\modified{DeepSeek-trained draft models are some exceptions in the community open-source models where temperature scaling gives limited gains, as shown in Table~\ref{tab:deepseek_summary}.
We can observe that the top-1 probability gaps of these models are generally significantly smaller compared to that presented in Figure~\ref{fig:matched_top1_prob_ent}, suggesting less underconfidence for posthoc calibration methods to correct. As discussed in Remark~\ref{rem:distribution_gap}, it would be an interesting future direction to find out which aspects in the training may explain this difference. More details of the results are in Appendix~\ref{appendix:observations}.
}

\subsection{Dirichlet Scaling}\label{sec:dirichlet_scaling}

As presented in the previous section, simple temperature scaling already shows significant improvement, which uses only one parameter to adjust the overall sharpness of the draft distribution. It is natural to consider whether there is room for even more gain with a stronger calibration parameter group.
In this section, we examine a family of Dirichlet scaling~\citep{kull2019beyond} methods.
Given this original probability distribution $p$, Dirichlet scaling seeks transformation parameters $W \in \RR^{V\times V}$ and $b \in \RR^V$ such that the adjusted distribution is
\begin{align}\label{eq:dirichlet_scaling}
    p_{\mathrm{cal}} = \mathrm{softmax}\left( W \log p + b\right) .
\end{align}
We consider $b=0$ in this paper. Dirichlet scaling is a more general calibration method class, recovering temperature scaling by taking $W = (1/T)I$. Beyond $W \propto I$, it provides more fine-grained scaling for each candidate token and allows candidate-interactive transformation with off-diagonal entries of $W$. \citet{kull2019beyond} presents evidence for improving confidence calibration compared to temperature scaling, especially for the classes that are not the top-1 candidate. This is beneficial for the regime of draft calibration, since the acceptance relies on the gap between the full distributions $\Norm{p-q}_1$.
In LLM applications, it is generally infeasible to apply the scaling directly because of the large vocabulary size $V$. Since usually several top candidates take up the majority of probability, we take $W = \begin{bmatrix}
    W^{(K)} & \\ & (1/T_{\mathrm{res}}) I ,
\end{bmatrix}$
where $W^{(K)} \in \RR^{K\times K}$ and apply the transformation to the top $K$ candidates (with the largest probability) and a scalar $T_{\mathrm{res}}$ for the other candidates. This form guarantees degeneration to temperature scaling, while reducing the cost.

\subsubsection{Trajectory replay for efficient verification.}\label{sec:trajectory_replay}

\begin{wrapfigure}[10]{r}{0.48\textwidth}
    \centering
    \vspace{-7pt}
    \includegraphics[width=\linewidth]{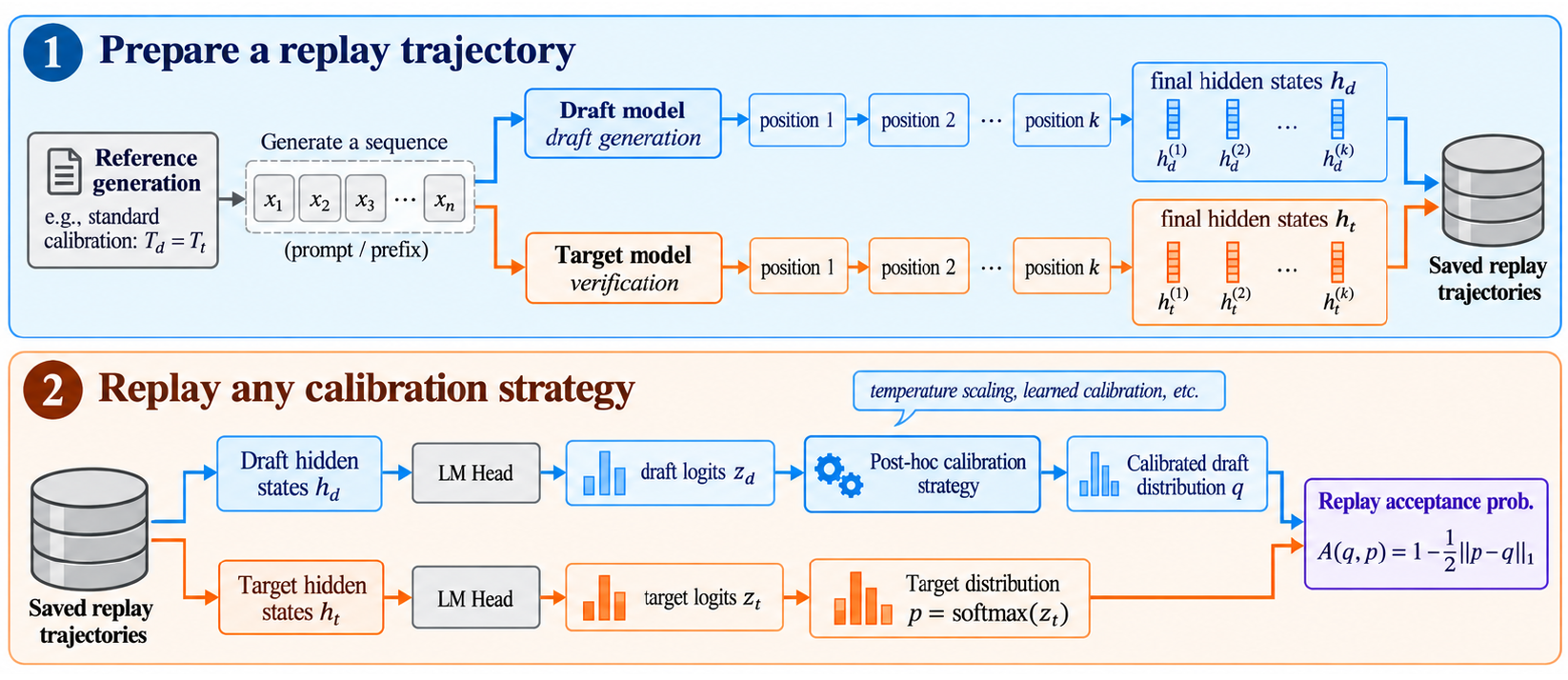}
    \par
    \setlength{\abovecaptionskip}{2pt}
    \caption{Illustration of the trajectory replay. 
    }
    \label{fig:replay_illustration}
\end{wrapfigure}

To test whether there is room for improvement by employing Dirichlet scaling efficiently, we introduce a trajectory replay pipeline to conduct quick checks for draft calibration methods. 
The idea of trajectory replay is \modified{to reuse fixed reference contexts, where} the acceptance probability can be computed by (\ref{eq:acceptance_probability}) given any $q$ and $p$ for a fixed draft position. Therefore, it is feasible to compute the performance of any draft calibration strategy on a saved specific trajectory together with the corresponding target and draft final hidden states -- directly saving logits costs too much space, so we save the final hidden states and recover the logits when replaying. 
\modified{Then, with the logits, we can compute the expected acceptance rate by $1 - \frac{1}{2} \Norm{p-q}_1$ and further record accepted length.}
The pipeline is illustrated in Figure~\ref{fig:replay_illustration}, and a detailed description is in Appendix~\ref{appendix:offline_replay}.

The efficiency benefit of the replay is huge: we conduct over a million replay evaluations
in the following sections within GPU minutes, while real evaluation may take millions of GPU hours.
There is indeed bias from this trajectory replay in simulating the speculative decoding performance in real inference. For example, we cannot straightforwardly transfer the definition of accepted length to trajectory replay, and we instead consider an alternative average accepted length $L_{\mathrm{pos}} := 1 + \sum_{i=1}^k \prod_{j=1}^i A_j$, where $A_j$ is the mean acceptance probability at draft position $j$ over eligible reference contexts.
Figure~\ref{fig:replay_online_gap} in Appendix~\ref{appendix:offline_replay} shows that $L_{\mathrm{pos}}$ approximately tracks online temperature trends, with mean absolute relative errors of 1.83\%--2.64\% across the three model pairs. Replay is therefore useful for screening calibration candidates, although it does not reproduce every temperature ranking or exact online gain.

\begin{table}[t]
\centering
\setlength{\tabcolsep}{5pt}
\caption{
\modified{Test replay accepted length} $L_{\mathrm{pos}}$ averaged across tasks.
Temperature and diagonal Dirichlet scaling selecting \modified{parameters per task on validation and on test are listed separately}.
Full-matrix Dirichlet scaling selects $K$ and a trained checkpoint (epoch $\geq 1$) on each task's validation split. \modified{Diagonal selection includes scalar fallback.}
}
\label{tab:dirichlet-scaling-results}
\vspace{0.1cm}
\begin{tabular}{@{}lccc@{}}
\toprule
Method
& \makecell{GPT-OSS-120B\\EAGLE-3}
& \makecell{GPT-OSS-120B\\DFlash}
& \makecell{Qwen3.6-35B-A3B\\DSpark} \\
\midrule
$T_d = 1$ & 3.0739 & 3.6345 & 4.0731 \\
\modified{Temperature (validation)}
& \modified{3.5085} & \modified{4.0751} & \modified{4.4825} \\
\modified{Diagonal Dirichlet (validation)}
& \modified{3.5084} & \modified{4.0752} & \modified{4.4815} \\
Full-matrix Dirichlet
& 3.4824 & 4.0563 & 4.4192 \\
\midrule
Optimal temperature \modified{(test oracle)}
& 3.5085 & 4.0751 & 4.4847 \\
Optimal diagonal Dirichlet \modified{(test oracle)}
& 3.5085 & 4.0753 & 4.4850 \\
\bottomrule
\end{tabular}\vspace{-0.3cm}

\end{table}

\subsubsection{Dirichlet scaling generally does not exceed temperature scaling}
We evaluate the following results on a trajectory replay with $T_d = T_t = 1$ as the reference generation. 
Recall the Dirichlet scaling formula~(\ref{eq:dirichlet_scaling}) and our simplification. 
We consider two cases to thoroughly examine the method:
\vspace{-0.2cm}
\begin{itemize}
    \item A diagonal exhaustive search: we take $W^{(K)}$ to be diagonal, equivalent to assigning separate temperatures to the top-$K$ candidates and the tail, with $W^{(K)}[i,i]=1/T_i$. Temperature scaling is recovered when all temperatures coincide. For $K=7$, we independently search each of the eight temperatures over the grid $\mathcal{G} = \{0.2,0.4,0.6,0.8,1,1.2\}$, giving $6^8=1,679,616$ candidates, and select the optimum per task on the \modified{validation split, retaining the test optimum as an oracle and listed separately}.
    \item Full-matrix training: for $K\in\{1,2,3,7\}$, we jointly optimize an unrestricted $W^{(K)}$ and $T_{\mathrm{res}}>0$ with Adam, minimizing a top-$K$-plus-tail total-variation surrogate across equally weighted tasks. Each task selects $K$ and a trained checkpoint using validation $L_{\mathrm{pos}}$, evaluated with full-vocabulary overlap. Appendix~\ref{appendix:replay_training} gives the training details.
\end{itemize}
\vspace{-0.2cm}
Table~\ref{tab:dirichlet-scaling-results} shows that exhaustive diagonal scaling offers negligible gains over temperature scaling, while full matrices trained with our surrogate loss perform even slightly worse. This reveals that temperature scaling may have captured the consistent distribution gap that applies to general tokens. \modified{Also note that scalar temperatures selected in validation nearly match the test optima on all tasks.}

\vspace{-0.2cm}

\subsection{Token-Adaptive Calibration Head}

\begin{table}[t]
\centering
\caption{Test $L_{\mathrm{pos}}$ averaged equally across tasks, in the same settings as Table~\ref{tab:dirichlet-scaling-results}. \modified{Scalar temperature is selected on validation, and the separately reported test oracle is} selected separately for each task on the test grid and shared across its tokens; token oracles select per position. Learned heads select hyperparameters, a trained checkpoint (epoch $\geq 1$), and binary thresholds on each task's validation split.}\label{tab:token-adaptive-temperature-scaling}
\vspace{0.1cm}
\begin{tabular}{lccc}
\toprule
Method &  \makecell{GPT-OSS-120B\\EAGLE-3}
& \makecell{GPT-OSS-120B\\DFlash}
& \makecell{Qwen3.6-35B-A3B\\DSpark} \\
\midrule
$T_d=1$ & 3.0739 & 3.6345 & 4.0731 \\
\modified{Temperature (validation, per task)} & \modified{3.5085} & \modified{4.0751} & \modified{4.4825} \\
Temperature head ($h_d$) & 3.5111 & 4.1091 & 4.4765 \\
Temperature head ($h_d,h_{t,\mathrm{prev}}$) & 3.5170 & 4.1091 & 4.4726 \\
Binary head ($h_d$) & 3.4031 & 4.0045 & 4.3281 \\
Binary head ($h_d,h_{t,\mathrm{prev}}$) & 3.4002 & 3.9998 & 4.3234 \\
\midrule
Optimal $T_d$ (\modified{test oracle,} per task) & 3.5085 & 4.0751 & 4.4847 \\
Token oracle ($\mathcal G$) & 4.0598 & 4.7628 & 4.9247 \\
Token oracle ($\{0,1\}$) & 3.9263 & 4.5943 & 4.7941 \\
\bottomrule
\end{tabular}
\vspace{-0.3cm}

\end{table}

The trials in Dirichlet scaling reveal that temperature scaling may have recovered most of the distribution gap we can reach with post-hoc calibration methods shared across tokens within each task. We consider another axis in this section: calibration can be adaptive to each specific draft token, with evidence from confidence calibration in LLMs that a token-wise calibration head may help with better performance in confidence calibration~\citep{xie2024calibrating}. We consider only token-adaptive temperature scaling in this section. The related results are presented in Table~\ref{tab:token-adaptive-temperature-scaling}, with detailed evaluation and training settings in Appendix~\ref{appendix:replay_training}. We summarize the methods we tried and the observations.

\vspace{-0.2cm}
\begin{itemize}
    \item \textbf{Token-wise oracle shows improvement room.} At each replay position, we select the temperature maximizing {$A(T)=1-\frac12\Norm{p-q_T}_1$} from the candidate temperature grid $\mathcal G$ or $\{0,1\}$, where $T=0$ denotes greedy drafting. As in Table~\ref{tab:token-adaptive-temperature-scaling}, these target-informed oracles substantially outperform the task-specific scalar optimum. The gap is even larger than that between {per-task scalar temperature scaling} and the original fixed $T_d = 1$, showing promising room for token-adaptive calibration when it can be done perfectly with temperature scaling. Moreover, the small $\{ 0,1 \}$ temperature grid also recovers {approximately 70\%--76\% of the gain from a full temperature search grid}, suggesting potential patterns in token-adaptive scaling.
    \item \textbf{Learned calibration heads generally fail to capture it.} We train a separate lightweight head per task to predict a positive temperature or a binary choice $T_d\in\{0,1\}$, using the draft state $h_d$ alone or concatenated with $h_{t,\mathrm{prev}}$, the last eligible target state cached from the preceding verification block, as the input, to obtain information from both draft and target models to better predict the temperature for the current draft token.
    The training settings are specified in Appendix~\ref{appendix:replay_training}. 
    We consider this $\{0,1\}$ grid choice because of its good oracle performance and potential simplicity in predicting a binary classification problem.
    We can observe that all the trained temperature heads remain near, or even slightly worse than, global temperature scaling, regardless of the binary classification and input design. This failure to capture the potential benefits of token-adaptive scaling suggests that more modifications may be needed beyond pure post-hoc calibration.
\end{itemize}
\vspace{-0.2cm}

\section{Conclusions}
In this paper, we study the draft calibration for speculative decoding, focusing on post-hoc methods adjusting distributions at inference time. We show that applying a standard temperature scaling to the draft distribution already significantly improves the acceptance performance. We further examine token-adaptive temperature scaling and find potential room for improvement if we do it perfectly, but the current temperature head still fails to capture the improvement.
One limitation of the work is that we have only studied post-hoc methods, but training methods, which have been extensively studied in the confidence calibration literature, may also be important for studying draft calibration, e.g., utilizing the potential room in token-adaptive temperature scaling or figuring out the exact reason for the consistent distribution gap. We consider this an interesting future direction.

\section*{Acknowledgment}
Tong Zhang was supported in part by NSF grants 2416897, 2505932, and 2601666 and ONR grant N000142512318. A portion of the computing resources used in this work was provided by the National Center for Supercomputing Applications (NCSA) through Delta (NSF OAC-2005572) and DeltaAI (NSF OAC-2320345), as well as by Illinois Computes and the NAIRR Pilot (NAIRR250157).

\bibliographystyle{plainnat}
\bibliography{ref}

@article{li2025eagle3,
  title={Eagle-3: Scaling up inference acceleration of large language models via training-time test},
  author={Li, Yuhui and Wei, Fangyun and Zhang, Chao and Zhang, Hongyang},
  journal={Advances in Neural Information Processing Systems},
  volume={38},
  pages={136737--136756},
  year={2025}
}

@inproceedings{li2024eagle2,
  title={Eagle-2: Faster inference of language models with dynamic draft trees},
  author={Li, Yuhui and Wei, Fangyun and Zhang, Chao and Zhang, Hongyang},
  booktitle={Proceedings of the 2024 conference on empirical methods in natural language processing},
  pages={7421--7432},
  year={2024}
}

@article{li2024eagle,
  title={Eagle: Speculative sampling requires rethinking feature uncertainty},
  author={Li, Yuhui and Wei, Fangyun and Zhang, Chao and Zhang, Hongyang},
  journal={arXiv preprint arXiv:2401.15077},
  year={2024}
}

@article{chen2026dflash,
  title={Dflash: Block diffusion for flash speculative decoding},
  author={Chen, Jian and Liang, Yesheng and Liu, Zhijian},
  journal={arXiv preprint arXiv:2602.06036},
  year={2026}
}

@article{cheng2026dspark,
  title={DSpark: Confidence-Scheduled Speculative Decoding with Semi-Autoregressive Generation},
  author={Cheng, Xin and Yu, Xingkai and Shao, Chenze and Li, Jiashi and Xiong, Yunfan and Qian, Yi and Zhu, Jiaqi and Ma, Shirong and Zhang, Xiaokang and Ye, Jiasheng and others},
  journal={arXiv preprint arXiv:2607.05147},
  year={2026}
}

@inproceedings{leviathan2023fast,
  title={Fast inference from transformers via speculative decoding},
  author={Leviathan, Yaniv and Kalman, Matan and Matias, Yossi},
  booktitle={International conference on machine learning},
  pages={19274--19286},
  year={2023},
  organization={PMLR}
}

@article{chen2023accelerating,
  title={Accelerating large language model decoding with speculative sampling},
  author={Chen, Charlie and Borgeaud, Sebastian and Irving, Geoffrey and Lespiau, Jean-Baptiste and Sifre, Laurent and Jumper, John},
  journal={arXiv preprint arXiv:2302.01318},
  year={2023}
}

@inproceedings{guo2017calibration,
  title={On calibration of modern neural networks},
  author={Guo, Chuan and Pleiss, Geoff and Sun, Yu and Weinberger, Kilian Q},
  booktitle={International conference on machine learning},
  pages={1321--1330},
  year={2017},
  organization={PMLR}
}

@article{cai2024medusa,
  title={Medusa: Simple llm inference acceleration framework with multiple decoding heads},
  author={Cai, Tianle and Li, Yuhong and Geng, Zhengyang and Peng, Hongwu and Lee, Jason D and Chen, Deming and Dao, Tri},
  journal={arXiv preprint arXiv:2401.10774},
  year={2024}
}

@article{kull2019beyond,
  title={Beyond temperature scaling: Obtaining well-calibrated multi-class probabilities with dirichlet calibration},
  author={Kull, Meelis and Perello Nieto, Miquel and K{\"a}ngsepp, Markus and Silva Filho, Telmo and Song, Hao and Flach, Peter},
  journal={Advances in neural information processing systems},
  volume={32},
  year={2019}
}

@article{eldenk2026attention,
  title={Attention Drift: What Autoregressive Speculative Decoding Models Learn},
  author={Eldenk, Do{\u{g}}a{\c{c}} and Mohapatra, Payal and Comlek, Yigitcan and Oktay, Kaan and Zhang, Hongyang and Xia, Stephen},
  journal={arXiv preprint arXiv:2605.09992},
  year={2026}
}

@inproceedings{xie2024calibrating,
  title={Calibrating language models with adaptive temperature scaling},
  author={Xie, Johnathan and Chen, Annie S and Lee, Yoonho and Mitchell, Eric and Finn, Chelsea},
  booktitle={Proceedings of the 2024 Conference on Empirical Methods in Natural Language Processing},
  pages={18128--18138},
  year={2024}
}

@article{li2026dartree,
  title={DARTree: Speculative Diffusion Decoding with Autoregressive Draft Trees},
  author={Li, Tianyi and Luo, Yaxin and Shang, Xinyi and Shen, Zhiqiang},
  journal={arXiv preprint arXiv:2608.13524},
  year={2026}
}

@article{liu2026speculative,
  title={Speculative Decoding: Performance or Illusion?},
  author={Liu, Xiaoxuan and Yu, Jiaxiang and Park, Jongseok and Stoica, Ion and Cheung, Alvin},
  journal={arXiv preprint arXiv:2601.11580},
  year={2026},
  url={https://arxiv.org/abs/2601.11580v2}
}

@article{cobbe2021gsm8k,
  title={Training Verifiers to Solve Math Word Problems},
  author={Cobbe, Karl and Kosaraju, Vineet and Bavarian, Mohammad and Chen, Mark and Jun, Heewoo and Kaiser, Lukasz and Plappert, Matthias and Tworek, Jerry and Hilton, Jacob and Nakano, Reiichiro and Hesse, Christopher and Schulman, John},
  journal={arXiv preprint arXiv:2110.14168},
  year={2021},
  url={https://arxiv.org/abs/2110.14168}
}

@article{lightman2023lets,
      title={Let's Verify Step by Step}, 
      author={Lightman, Hunter and Kosaraju, Vineet and Burda, Yura and Edwards, Harri and Baker, Bowen and Lee, Teddy and Leike, Jan and Schulman, John and Sutskever, Ilya and Cobbe, Karl},
      journal={arXiv preprint arXiv:2305.20050},
      year={2023},
  url={https://arxiv.org/abs/2305.20050}
}

@article{dekoninck2026matharena,
      title={Beyond Benchmarks: {MathArena} as an Evaluation Platform for Mathematics with {LLMs}}, 
      author={Jasper Dekoninck and Nikola Jovanovi{\'c} and Tim Gehrunger and K{\'a}ri R{\"o}gnvaldsson and Ivo Petrov and Chenhao Sun and Martin Vechev},
      journal={arXiv preprint arXiv:2605.00674},
      year={2026},
      eprint={2605.00674},
      archivePrefix={arXiv},
      primaryClass={cs.CL},
      url={https://arxiv.org/abs/2605.00674}, 
}

@misc{chen2021evaluating,
      title={Evaluating Large Language Models Trained on Code},
      author={Mark Chen and Jerry Tworek and Heewoo Jun and Qiming Yuan and Henrique Ponde de Oliveira Pinto and Jared Kaplan and Harri Edwards and Yuri Burda and Nicholas Joseph and Greg Brockman and Alex Ray and Raul Puri and Gretchen Krueger and Michael Petrov and Heidy Khlaaf and Girish Sastry and Pamela Mishkin and Brooke Chan and Scott Gray and Nick Ryder and Mikhail Pavlov and Alethea Power and Lukasz Kaiser and Mohammad Bavarian and Clemens Winter and Philippe Tillet and Felipe Petroski Such and Dave Cummings and Matthias Plappert and Fotios Chantzis and Elizabeth Barnes and Ariel Herbert-Voss and William Hebgen Guss and Alex Nichol and Alex Paino and Nikolas Tezak and Jie Tang and Igor Babuschkin and Suchir Balaji and Shantanu Jain and William Saunders and Christopher Hesse and Andrew N. Carr and Jan Leike and Josh Achiam and Vedant Misra and Evan Morikawa and Alec Radford and Matthew Knight and Miles Brundage and Mira Murati and Katie Mayer and Peter Welinder and Bob McGrew and Dario Amodei and Sam McCandlish and Ilya Sutskever and Wojciech Zaremba},
      year={2021},
      eprint={2107.03374},
      archivePrefix={arXiv},
      primaryClass={cs.LG},
  url={https://arxiv.org/abs/2107.03374}
}

@article{austin2021program,
  title={Program Synthesis with Large Language Models},
  author={Austin, Jacob and Odena, Augustus and Nye, Maxwell and Bosma, Maarten and Michalewski, Henryk and Dohan, David and Jiang, Ellen and Cai, Carrie and Terry, Michael and Le, Quoc and others},
  journal={arXiv preprint arXiv:2108.07732},
  year={2021},
  url={https://arxiv.org/abs/2108.07732}
}

@article{jain2024livecodebench,
  author    = {Naman Jain and King Han and Alex Gu and Wen-Ding Li and Fanjia Yan and Tianjun Zhang and Sida Wang and Armando Solar-Lezama and Koushik Sen and Ion Stoica},
  title     = {{LiveCodeBench}: Holistic and Contamination Free Evaluation of Large Language Models for Code},
  year      = {2024},
  journal   = {arXiv preprint arXiv:2403.07974},
  url={https://arxiv.org/abs/2403.07974}
}

@misc{zheng2023judging,
      title={Judging {LLM}-as-a-judge with {MT-Bench} and {Chatbot Arena}}, 
      author={Lianmin Zheng and Wei-Lin Chiang and Ying Sheng and Siyuan Zhuang and Zhanghao Wu and Yonghao Zhuang and Zi Lin and Zhuohan Li and Dacheng Li and Eric P. Xing and Hao Zhang and Joseph E. Gonzalez and Ion Stoica},
      year={2023},
      eprint={2306.05685},
      archivePrefix={arXiv},
      primaryClass={cs.CL},
  url={https://arxiv.org/abs/2306.05685}
}

@article{li2024crowdsourced,
  title={From Crowdsourced Data to High-Quality Benchmarks: {Arena-Hard} and {BenchBuilder} Pipeline},
  author={Li, Tianle and Chiang, Wei-Lin and Frick, Evan and Dunlap, Lisa and Wu, Tianhao and Zhu, Banghua and Gonzalez, Joseph E and Stoica, Ion},
  journal={arXiv preprint arXiv:2406.11939},
  year={2024},
  url={https://arxiv.org/abs/2406.11939}
}

@article{li2026specforge,
  title={{SpecForge}: A Flexible and Efficient Open-Source Training Framework for Speculative Decoding},
  author={Li, Shenggui and Wang, Chao and Zhu, Yikai and Wang, Yubo and Yin, Fan and Shi, Shuai and Chen, Yefei and Dong, Xiaomin and Chen, Qiaoling and Pan, Jin and Li, Ji and Xie, Laixin and Zhang, Yineng and Yu, Lei and Wen, Yonggang and Tsang, Ivor and Zhang, Tianwei},
  journal={arXiv preprint arXiv:2603.18567},
  year={2026},
  url={https://arxiv.org/abs/2603.18567}
}

@inproceedings{zheng2024sglang,
  title={{SGLang}: Efficient Execution of Structured Language Model Programs},
  author={Zheng, Lianmin and Yin, Liangsheng and Xie, Zhiqiang and Sun, Chuyue and Huang, Jeff and Yu, Cody Hao and Cao, Shiyi and Kozyrakis, Christos and Stoica, Ion and Gonzalez, Joseph E. and Barrett, Clark and Sheng, Ying},
  booktitle={Advances in Neural Information Processing Systems},
  volume={37},
  pages={62557--62583},
  year={2024},
  doi={10.52202/079017-2000},
  url={https://proceedings.neurips.cc/paper_files/paper/2024/file/724be4472168f31ba1c9ac630f15dec8-Paper-Conference.pdf}
}

@inproceedings{zhou2024distillspec,
 author = {Zhou, Yongchao and Lyu, Kaifeng and Rawat, Ankit Singh and Menon, Aditya Krishna and Rostamizadeh, Afshin and Kumar, Sanjiv and Kagy, Jean-Fran\c{c}ois and Agarwal, Rishabh},
 booktitle = {International Conference on Learning Representations},
 editor = {B. Kim and Y. Yue and S. Chaudhuri and K. Fragkiadaki and M. Khan and Y. Sun},
 pages = {32011--32050},
 title = {{DistillSpec}: Improving Speculative Decoding via Knowledge Distillation},
 url = {https://proceedings.iclr.cc/paper_files/paper/2024/file/8766fbc68e1ed1cdef712ce273e0a363-Paper-Conference.pdf},
 volume = {2024},
 year = {2024}
}

@inproceedings{desai2020calibration,
    title = "Calibration of Pre-trained Transformers",
    author = "Desai, Shrey  and
      Durrett, Greg",
    editor = "Webber, Bonnie  and
      Cohn, Trevor  and
      He, Yulan  and
      Liu, Yang",
    booktitle = "Proceedings of the 2020 Conference on Empirical Methods in Natural Language Processing (EMNLP)",
    month = nov,
    year = "2020",
    address = "Online",
    publisher = "Association for Computational Linguistics",
    url = "https://aclanthology.org/2020.emnlp-main.21/",
    doi = "10.18653/v1/2020.emnlp-main.21",
    pages = "295--302"
}

@inproceedings{ouyang2024temperature,
    title = "Temperature-Centric Investigation of Speculative Decoding with Knowledge Distillation",
    author = "Ouyang, Siru  and
      Wang, Shuohang  and
      Jiang, Minhao  and
      Zhong, Ming  and
      Yu, Donghan  and
      Han, Jiawei  and
      Shen, Yelong",
    editor = "Al-Onaizan, Yaser  and
      Bansal, Mohit  and
      Chen, Yun-Nung",
    booktitle = "Findings of the Association for Computational Linguistics: EMNLP 2024",
    month = nov,
    year = "2024",
    address = "Miami, Florida, USA",
    publisher = "Association for Computational Linguistics",
    url = "https://aclanthology.org/2024.findings-emnlp.767/",
    doi = "10.18653/v1/2024.findings-emnlp.767",
    pages = "13125--13137"
}

@inproceedings{wang2025subspec,
 author = {Wang, Pei-Shuo and Chen, Jian-Jia and Yang, Chun-Che and Chang, Chi-Chih and Huang, Ning-Chi and Abdelfattah, Mohamed S. and Wu, Kai-Chiang},
 booktitle = {Advances in Neural Information Processing Systems},
 doi = {10.52202/085713-1815},
 editor = {D. Belgrave and C. Zhang and H. Lin and R. Pascanu and P. Koniusz and M. Ghassemi and N. Chen},
 pages = {54334--54357},
 publisher = {Curran Associates, Inc.},
 title = {Speculate Deep and Accurate: Lossless and Training-Free Acceleration for Offloaded {LLMs} via Substitute Speculative Decoding},
 url = {https://proceedings.neurips.cc/paper_files/paper/2025/file/4e32ca54ae8a3a0ea4c1fd3663f73b0f-Paper-Conference.pdf},
 volume = {38},
 year = {2025}
}

@inproceedings{zhou2026calibrated,
    title = "Calibrated Speculative Decoding: Frequency-Guided Candidate Selection for Efficient Inference",
    author = "Zhou, Xuwen  and
      Liu, Fangxin  and
      Wang, Chao  and
      Zheng, Xiao  and
      Zheng, Hao  and
      He, Min  and
      Jiang, Li  and
      Guan, Haibing",
    editor = "Liakata, Maria  and
      Moreira, Viviane P.  and
      Zhang, Jiajun  and
      Jurgens, David",
    booktitle = "Proceedings of the 64th Annual Meeting of the {A}ssociation for {C}omputational {L}inguistics (Volume 1: Long Papers)",
    month = jul,
    year = "2026",
    address = "San Diego, California, United States",
    publisher = "Association for Computational Linguistics",
    url = "https://aclanthology.org/2026.acl-long.1369/",
    doi = "10.18653/v1/2026.acl-long.1369",
    pages = "29677--29688",
    ISBN = "979-8-89176-390-6"
}

@inproceedings{choi2025mamba,
    title = "{Mamba} Drafters for Speculative Decoding",
    author = "Choi, Daewon  and
      Oh, Seunghyuk  and
      Dingliwal, Saket  and
      Tack, Jihoon  and
      Kim, Kyuyoung  and
      Song, Woomin  and
      Kim, Seojin  and
      Han, Insu  and
      Shin, Jinwoo  and
      Galstyan, Aram  and
      Katiyar, Shubham  and
      Bodapati, Sravan Babu",
    editor = "Christodoulopoulos, Christos  and
      Chakraborty, Tanmoy  and
      Rose, Carolyn  and
      Peng, Violet",
    booktitle = "Findings of the Association for Computational Linguistics: EMNLP 2025",
    month = nov,
    year = "2025",
    address = "Suzhou, China",
    publisher = "Association for Computational Linguistics",
    url = "https://aclanthology.org/2025.findings-emnlp.1167/",
    doi = "10.18653/v1/2025.findings-emnlp.1167",
    pages = "21404--21418",
    ISBN = "979-8-89176-335-7"
}

@article{xiao2024parallelspec,
  title={{ParallelSpec}: Parallel Drafter for Efficient Speculative Decoding},
  author={Xiao, Zilin and Zhang, Hongming and Ge, Tao and Ouyang, Siru and Ordonez, Vicente and Yu, Dong},
  journal={arXiv preprint arXiv:2410.05589},
  year={2024},
  url={https://arxiv.org/abs/2410.05589}
}

@article{hui2026peagle,
  title={{P-EAGLE}: Parallel-Drafting {EAGLE} with Scalable Training},
  author={Hui, Mude and Huang, Xin and Salas, Jaime Campos and Sun, Yue and Pemberton, Nathan and Song, Xiang and Khetan, Ashish and Karypis, George},
  journal={arXiv preprint arXiv:2602.01469},
  year={2026},
  url={https://arxiv.org/abs/2602.01469}
}

@article{zhang2026flexdraft,
  title={{FlexDraft}: Flexible Speculative Decoding via Attention Tuning and Bonus-Guided Calibration},
  author={Zhang, Yaojie and Huang, Jianuo and Ke, Junlong and Han, Yuhang and Long, Yongji and Zhao, Tianchen and Qi, Biqing and Zhang, Linfeng},
  journal={arXiv preprint arXiv:2605.20022},
  year={2026},
  url={https://arxiv.org/abs/2605.20022}
}

@article{wen2026specbound,
  title={{SpecBound}: Adaptive Bounded Self-Speculation with Layer-wise Confidence Calibration},
  author={Wen, Zhuofan and Feng, Yang},
  journal={arXiv preprint arXiv:2604.12247},
  year={2026},
  url={https://arxiv.org/abs/2604.12247}
}

@inproceedings{miao2024specinfer,
  title={{SpecInfer}: Accelerating Large Language Model Serving with Tree-based Speculative Inference and Verification},
  author={Miao, Xupeng and Oliaro, Gabriele and Zhang, Zhihao and Cheng, Xinhao and Wang, Zeyu and Zhang, Zhengxin and Wong, Rae Ying Yee and Zhu, Alan and Yang, Lijie and Shi, Xiaoxiang and Shi, Chunan and Chen, Zhuoming and Arfeen, Daiyaan and Abhyankar, Reyna and Jia, Zhihao},
  booktitle={Proceedings of the 29th ACM International Conference on Architectural Support for Programming Languages and Operating Systems, Volume 3},
  year={2024},
  doi={10.1145/3620666.3651335},
  url={https://arxiv.org/abs/2305.09781v4}
}

@inproceedings{stern2018blockwise,
 author = {Stern, Mitchell and Shazeer, Noam and Uszkoreit, Jakob},
 booktitle = {Advances in Neural Information Processing Systems},
 publisher = {Curran Associates, Inc.},
 title = {Blockwise Parallel Decoding for Deep Autoregressive Models},
 url = {https://proceedings.neurips.cc/paper_files/paper/2018/file/c4127b9194fe8562c64dc0f5bf2c93bc-Paper.pdf},
 volume = {31},
 year = {2018}
}

@inproceedings{xia2023speculative,
    title = "Speculative Decoding: Exploiting Speculative Execution for Accelerating Seq2seq Generation",
    author = "Xia, Heming  and
      Ge, Tao  and
      Wang, Peiyi  and
      Chen, Si-Qing  and
      Wei, Furu  and
      Sui, Zhifang",
    booktitle = "Findings of the Association for Computational Linguistics: EMNLP 2023",
    month = dec,
    year = "2023",
    address = "Singapore",
    publisher = "Association for Computational Linguistics",
    url = "https://aclanthology.org/2023.findings-emnlp.257/",
    doi = "10.18653/v1/2023.findings-emnlp.257",
    pages = "3909--3925"
}

@inproceedings{zhang2024draftverify,
    title = "Draft {\&} Verify: Lossless Large Language Model Acceleration via Self-Speculative Decoding",
    author = "Zhang, Jun  and
      Wang, Jue  and
      Li, Huan  and
      Shou, Lidan  and
      Chen, Ke  and
      Chen, Gang  and
      Mehrotra, Sharad",
    booktitle = "Proceedings of the 62nd Annual Meeting of the Association for Computational Linguistics (Volume 1: Long Papers)",
    month = aug,
    year = "2024",
    address = "Bangkok, Thailand",
    publisher = "Association for Computational Linguistics",
    url = "https://aclanthology.org/2024.acl-long.607/",
    doi = "10.18653/v1/2024.acl-long.607",
    pages = "11263--11282"
}

@inproceedings{liu2024kangaroo,
 author = {Liu, Fangcheng and Tang, Yehui and Liu, Zhenhua and Ni, Yunsheng and Tang, Duyu and Han, Kai and Wang, Yunhe},
 booktitle = {Advances in Neural Information Processing Systems},
 doi = {10.52202/079017-0381},
 pages = {11946--11965},
 publisher = {Curran Associates, Inc.},
 title = {{Kangaroo}: Lossless Self-Speculative Decoding for Accelerating {LLMs} via Double Early Exiting},
 url = {https://proceedings.neurips.cc/paper_files/paper/2024/file/16336d94a5ffca8de019087ab7fe403f-Paper-Conference.pdf},
 volume = {37},
 year = {2024}
}

@inproceedings{ankner2024hydra,
  title={{Hydra}: Sequentially-Dependent Draft Heads for {Medusa} Decoding},
  author={Ankner, Zachary and Parthasarathy, Rishab and Nrusimha, Aniruddha and Rinard, Christopher and Ragan-Kelley, Jonathan and Brandon, William},
  booktitle={First Conference on Language Modeling},
  year={2024},
  url={https://arxiv.org/abs/2402.05109}
}

@inproceedings{zhang2025hass,
  title={Learning Harmonized Representations for Speculative Sampling},
  author={Zhang, Lefan and Wang, Xiaodan and Huang, Yanhua and Xu, Ruiwen},
  booktitle={International Conference on Learning Representations},
  year={2025},
  url={https://arxiv.org/abs/2408.15766}
}

@inproceedings{an2026pard,
  title={{PARD}: Accelerating {LLM} Inference with Low-Cost {PAR}allel Draft Model Adaptation},
  author={An, Zihao and Bai, Huajun and Liu, Ziqiong and Li, Dong and Barsoum, Emad},
  booktitle={International Conference on Learning Representations},
  year={2026},
  url={https://iclr.cc/virtual/2026/poster/10008956}
}

@article{wang2023calibration,
  title={Calibration in Deep Learning: A Survey of the State-of-the-Art},
  author={Wang, Cheng},
  journal={arXiv preprint arXiv:2308.01222},
  year={2023},
  url={https://arxiv.org/abs/2308.01222}
}

@inproceedings{naeini2015obtaining,
  title={Obtaining Well Calibrated Probabilities Using Bayesian Binning},
  author={Naeini, Mahdi Pakdaman and Cooper, Gregory F. and Hauskrecht, Milos},
  booktitle={Proceedings of the AAAI Conference on Artificial Intelligence},
  volume={29},
  year={2015},
  pages={2901--2907},
  doi={10.1609/aaai.v29i1.9602},
  url={https://ojs.aaai.org/index.php/AAAI/article/view/9602}
}

@InProceedings{kull2017beta,
  title = 	 {{Beta calibration: a well-founded and easily implemented improvement on logistic calibration for binary classifiers}},
  author = 	 {Kull, Meelis and Filho, Telmo Silva and Flach, Peter},
  booktitle = 	 {Proceedings of the 20th International Conference on Artificial Intelligence and Statistics},
  pages = 	 {623--631},
  year = 	 {2017},
  volume = 	 {54},
  series = 	 {Proceedings of Machine Learning Research},
  month = 	 {20--22 Apr},
  publisher =    {PMLR},
  url = 	 {https://proceedings.mlr.press/v54/kull17a.html},
}

@inproceedings{kumar2019verified,
 author = {Kumar, Ananya and Liang, Percy S and Ma, Tengyu},
 booktitle = {Advances in Neural Information Processing Systems},
 publisher = {Curran Associates, Inc.},
 title = {Verified Uncertainty Calibration},
 url = {https://proceedings.neurips.cc/paper_files/paper/2019/file/f8c0c968632845cd133308b1a494967f-Paper.pdf},
 volume = {32},
 year = {2019}
}

@InProceedings{zhang2020mix,
  title = 	 {{Mix-n-Match}: Ensemble and Compositional Methods for Uncertainty Calibration in Deep Learning},
  author =       {Zhang, Jize and Kailkhura, Bhavya and Han, T. Yong-Jin},
  booktitle = 	 {Proceedings of the 37th International Conference on Machine Learning},
  pages = 	 {11117--11128},
  year = 	 {2020},
  volume = 	 {119},
  series = 	 {Proceedings of Machine Learning Research},
  month = 	 {13--18 Jul},
  publisher =    {PMLR},
  url = 	 {https://proceedings.mlr.press/v119/zhang20k.html},
}

@InProceedings{ding2021local,
    author    = {Ding, Zhipeng and Han, Xu and Liu, Peirong and Niethammer, Marc},
    title     = {Local Temperature Scaling for Probability Calibration},
    booktitle = {Proceedings of the IEEE/CVF International Conference on Computer Vision (ICCV)},
    month     = {October},
    year      = {2021},
    pages     = {6889-6899},
    url = {https://openaccess.thecvf.com/content/ICCV2021/html/Ding_Local_Temperature_Scaling_for_Probability_Calibration_ICCV_2021_paper.html}
}

@InProceedings{kumar2018trainable,
  title = 	 {Trainable Calibration Measures for Neural Networks from Kernel Mean Embeddings},
  author =       {Kumar, Aviral and Sarawagi, Sunita and Jain, Ujjwal},
  booktitle = 	 {Proceedings of the 35th International Conference on Machine Learning},
  pages = 	 {2805--2814},
  year = 	 {2018},
  volume = 	 {80},
  series = 	 {Proceedings of Machine Learning Research},
  month = 	 {10--15 Jul},
  publisher =    {PMLR},
  url = 	 {https://proceedings.mlr.press/v80/kumar18a.html},
}

@inproceedings{mukhoti2020calibrating,
 author = {Mukhoti, Jishnu and Kulharia, Viveka and Sanyal, Amartya and Golodetz, Stuart and Torr, Philip H. S. and Dokania, Puneet K.},
 booktitle = {Advances in Neural Information Processing Systems},
 pages = {15288--15299},
 publisher = {Curran Associates, Inc.},
 title = {Calibrating Deep Neural Networks using Focal Loss},
 url = {https://proceedings.neurips.cc/paper_files/paper/2020/file/aeb7b30ef1d024a76f21a1d40e30c302-Paper.pdf},
 volume = {33},
 year = {2020}
}

@inproceedings{berdoz2026steering,
  title={Steering Pretrained Drafters During Speculative Decoding},
  author={Berdoz, Fr{\'e}d{\'e}ric and Rheinboldt, Peer and Wattenhofer, Roger},
  booktitle={Proceedings of the AAAI Conference on Artificial Intelligence},
  volume={40},
  year={2026},
  pages={30067--30075},
  doi={10.1609/aaai.v40i36.40255},
  url={https://ojs.aaai.org/index.php/AAAI/article/view/40255}
}

@article{samarin2026lk,
  title={LK losses: Direct acceptance rate optimization for speculative decoding},
  author={Samarin, Alexander and Krutikov, Sergei and Shevtsov, Anton and Skvortsov, Sergei and Fisin, Filipp and Golubev, Alexander},
  journal={arXiv preprint arXiv:2602.23881},
  year={2026}
}

\appendix

\section{Detailed Experiment Settings}\label{appendix:experiment_settings}

\subsection{Models and decoding configuration}
Table~\ref{tab:model_settings} lists all nine model pairs in Figure~\ref{fig:temperature_scaling_accepted_length}. Here $K$ counts proposed tokens, excluding the additional correction or bonus token, so a fully accepted block commits $K+1$ tokens. Chain refers to the proposal topology: EAGLE-3 generates proposals autoregressively, while DFlash and DSpark use their block drafting mechanisms without branching. We keep the released model weights fixed and apply calibration at inference time. 
We did not apply the confidence-based verification truncation for DSpark to keep the budget fixed.

\begin{table}[H]
\centering
\caption{Target and draft checkpoints used in the experiments. Target names and draft repository IDs link to their Hugging Face pages. $K$ is the number of draft positions, fixed within each model pair.}
\label{tab:model_settings}
\begingroup
\footnotesize
\setlength{\tabcolsep}{4pt}
\renewcommand{\arraystretch}{1.15}
\def\UrlBreaks{\do\/\do\-\do\.}
\begin{tabular}{@{}>{\raggedright\arraybackslash}p{0.27\textwidth}lc >{\raggedright\arraybackslash}p{0.50\textwidth}@{}}
\toprule
Target model & Method & $K$ & Draft checkpoint \\
\midrule
\href{https://huggingface.co/google/gemma-4-31B-it}{Gemma-4-31B-IT} & EAGLE-3 & 5 & \href{https://huggingface.co/RedHatAI/gemma-4-31B-it-speculator.eagle3}{\nolinkurl{RedHatAI/gemma-4-31B-it-speculator.eagle3}} \\
\href{https://huggingface.co/Qwen/Qwen3-30B-A3B-Instruct-2507}{Qwen3-30B-A3B-Instruct-2507} & EAGLE-3 & 3 & \href{https://huggingface.co/lmsys/SGLang-EAGLE3-Qwen3-30B-A3B-Instruct-2507-SpecForge-Nex}{\nolinkurl{lmsys/SGLang-EAGLE3-Qwen3-30B-A3B-Instruct-2507-SpecForge-Nex}} \\
\href{https://huggingface.co/openai/gpt-oss-120b}{GPT-OSS-120B} & EAGLE-3 & 9 & \href{https://huggingface.co/lmsys/EAGLE3-gpt-oss-120b-bf16}{\nolinkurl{lmsys/EAGLE3-gpt-oss-120b-bf16}} \\
\midrule
\href{https://huggingface.co/Qwen/Qwen3.5-27B}{Qwen3.5-27B} & DFlash & 7 & \href{https://huggingface.co/z-lab/Qwen3.5-27B-DFlash}{\nolinkurl{z-lab/Qwen3.5-27B-DFlash}} \\
\href{https://huggingface.co/Qwen/Qwen3.6-35B-A3B}{Qwen3.6-35B-A3B} & DFlash & 15 & \href{https://huggingface.co/z-lab/Qwen3.6-35B-A3B-DFlash}{\nolinkurl{z-lab/Qwen3.6-35B-A3B-DFlash}} \\
\href{https://huggingface.co/openai/gpt-oss-120b}{GPT-OSS-120B} & DFlash & 9 & \href{https://huggingface.co/z-lab/gpt-oss-120b-DFlash}{\nolinkurl{z-lab/gpt-oss-120b-DFlash}} \\
\midrule
\href{https://huggingface.co/google/gemma-4-31B-it}{Gemma-4-31B-IT} & DSpark & 7 & \href{https://huggingface.co/RedHatAI/gemma-4-31B-it-speculator.dspark}{\nolinkurl{RedHatAI/gemma-4-31B-it-speculator.dspark}} \\
\href{https://huggingface.co/Qwen/Qwen3.6-35B-A3B}{Qwen3.6-35B-A3B} & DSpark & 8 & \href{https://huggingface.co/RedHatAI/Qwen3.6-35B-A3B-speculator.dspark}{\nolinkurl{RedHatAI/Qwen3.6-35B-A3B-speculator.dspark}} \\
\href{https://huggingface.co/Qwen/Qwen3.8-27B-FP8}{Qwen3.8-27B-FP8} & DSpark & 7 & \href{https://huggingface.co/RadixArk/Qwen3.8-27B-DSpark}{\nolinkurl{RadixArk/Qwen3.8-27B-DSpark}} \\
\bottomrule
\end{tabular}
\endgroup
\end{table}

\paragraph{Sampling and prompt variants.}
Temperature scaling sweeps use $T_d\in\{0,0.2,0.4,0.6,0.8,1,1.2\}$ with $T_t=1$; $T_d=0$ denotes a point mass at the draft argmax. We disable sampling truncation (\texttt{top\_p=1}, \texttt{top\_k=-1}) and use base seed 980406 with request-index offsets. Chat runs use the target tokenizer's chat template. Completion uses the same task prompt without a chat template. 
The distribution plots in Figures~\ref{fig:topk_set_overlap} and~\ref{fig:matched_top1_prob_ent} use the low/non-thinking settings. The calibration replay uses the low/non-thinking settings. Best temperatures in the online plots are selected retrospectively from the displayed grid.

\paragraph{Serving configuration.}
We evaluate acceptance using our SpecForge/SGLang serving integration with independent draft-temperature control and probability-ratio rejection sampling. The per-server active-request limits in the nine panels of Figure~\ref{fig:temperature_scaling_accepted_length}, in row-major order, are 8, 64, 64, 8, 16, 64, 16, 20, and 18. Client concurrency can exceed these limits because requests queue; these values are capacity limits under continuous batching, not fixed batch sizes. Comparisons vary calibration within a model pair and retain its serving configuration, rather than comparing throughput across different model pairs.

\subsection{Datasets}
\begin{table}[H]
\centering
\caption{Evaluation datasets, broad task categories, and numbers of prompts used per setting. Repository links identify the Hugging Face sources; the text specifies the frozen evaluation snapshot.}
\label{tab:dataset_settings}
\begingroup
\footnotesize
\setlength{\tabcolsep}{4pt}
\renewcommand{\arraystretch}{1.15}
\def\UrlBreaks{\do\/\do\-\do\.\do\_}
\begin{tabular}{@{}l>{\raggedright\arraybackslash}p{0.36\textwidth}r>{\raggedright\arraybackslash}p{0.43\textwidth}@{}}
\toprule
Domain & Benchmark & Prompts & Hugging Face dataset \\
\midrule
Math & GSM8K~\citep{cobbe2021gsm8k} & 1,319 & \href{https://huggingface.co/datasets/openai/gsm8k}{\nolinkurl{openai/gsm8k}} \\
Math & MATH-500~\citep{lightman2023lets} & 500 & \href{https://huggingface.co/datasets/HuggingFaceH4/MATH-500}{\nolinkurl{HuggingFaceH4/MATH-500}} \\
Math & AIME 2025~\citep{dekoninck2026matharena} & 30 & \href{https://huggingface.co/datasets/MathArena/aime_2025}{\nolinkurl{MathArena/aime_2025}} \\
Code & HumanEval~\citep{chen2021evaluating} & 164 & \href{https://huggingface.co/datasets/openai/openai_humaneval}{\nolinkurl{openai/openai_humaneval}} \\
Code & MBPP~\citep{austin2021program} & 257 & \href{https://huggingface.co/datasets/google-research-datasets/mbpp}{\nolinkurl{google-research-datasets/mbpp}} \\
Code & LiveCodeBench~\citep{jain2024livecodebench} & 1,055 & \href{https://huggingface.co/datasets/livecodebench/code_generation_lite}{\nolinkurl{livecodebench/code_generation_lite}} \\
Chat & MT-Bench~\citep{zheng2023judging} & 80 & \href{https://huggingface.co/datasets/HuggingFaceH4/mt_bench_prompts}{\nolinkurl{HuggingFaceH4/mt_bench_prompts}} \\
Chat & Arena-Hard v2~\citep{li2024crowdsourced} & 750 & \href{https://huggingface.co/datasets/lmarena-ai/arena-hard-auto}{\nolinkurl{lmarena-ai/arena-hard-auto}} \\
\midrule
& Total & 4,155 & \\
\bottomrule
\end{tabular}
\endgroup
\end{table}

We use the frozen prompt JSONL files from \href{https://github.com/deepseek-ai/DeepSpec/tree/005e03b81cec38b7da6399833d609ee89a2587f2/eval_datasets}{DeepSpec's evaluation suite at revision \texttt{005e03b8}}, retaining all rows of the eight datasets in Table~\ref{tab:dataset_settings}. These comprise GSM8K's \texttt{main/test} split, MATH-500's \texttt{test} split, MathArena AIME 2025's \texttt{train} split, HumanEval's \texttt{test} split, MBPP's \texttt{sanitized/test} split, the 1,055-problem LiveCodeBench snapshot assembled from \texttt{test.jsonl} through \texttt{test6.jsonl}, MT-Bench's 80 conversations, and Arena-Hard's \texttt{data/arena-hard-v2.0/question.jsonl}. Split names here follow the dataset repositories; all serve as prompt workloads for our decoding experiments.

\subsection{Replay evaluation and calibration training}\label{appendix:replay_training}
The replay experiments use GPT-OSS-120B/EAGLE-3 and GPT-OSS-120B/DFlash with low reasoning effort ($K=9$), and Qwen3.6-35B-A3B/DSpark with thinking disabled ($K=8$). Reference trajectories use $T_d=T_t=1$. We cache final hidden states in BF16 and reconstruct distributions with the corresponding output heads, including model-specific normalization and vocabulary mappings. Eligible positions are determined by acceptance of all preceding proposals on the reference trajectory. Within each task, we split complete requests into 40\%/10\%/50\% train/validation/test partitions; all blocks from a request stay together. This yields 1,660/413/2,082 requests per model pair. All calibration tables use the same test partition and average tasks equally. Replay scores describe these fixed contexts and do not directly measure fresh-rollout throughput.

\paragraph{Dirichlet calibration.}
The diagonal search uses the six-point temperature grid $\{0.2,0.4,0.6,0.8,1,1.2\}$ independently for each of the top seven ranks and the tail, giving $6^8=1,679,616$ candidates. Full-matrix calibration uses $K\in\{1,2,3,7\}$ calibrated vocabulary ranks, independently of the draft chain length. The unrestricted $W^{(K)}$ acts on ranks sorted by the original draft probabilities, and $T_{\mathrm{res}}=\exp(-s)>0$ is jointly trained, with no additive bias. Each supported tail token retains its individual transformed score before full-vocabulary normalization.

Let $v_1,\ldots,v_K$ be the draft's top-$K$ token IDs and $g_K(r)=(r(v_1),\ldots,r(v_K),1-\sum_{j=1}^K r(v_j))$. We minimize the coarse total-variation loss $\ell_{\mathrm{TV}}=\frac12\lVert g_K(q_{\mathrm{cal}})-g_K(p)\rVert_1$. The target residual includes mass outside the draft vocabulary; grouping is used only for the loss. This surrogate lower-bounds full-distribution total variation, whereas validation and test scores use exact full-vocabulary overlap. Matrices are trained across tasks with equal task weight; Table~\ref{tab:dirichlet-scaling-results} selects the rank cutoff, learning rate, initialization, and checkpoint separately for each task using validation $L_{\mathrm{pos}}$. There is no separate matrix training per task. Temperature and diagonal \modified{deployment choices are selected per task on validation, and test-selected optima are retained as oracle comparators}. \modified{We retain the scalar if its validation score is at least as high as the selected diagonal candidate. }

\paragraph{Token-adaptive heads.}
We train a separate head per task, using either the final normalized draft state $h_d$ or its concatenation with the last eligible target final-layer state cached from the preceding verification block, $h_{t,\mathrm{prev}}$. Each input vector is RMS-normalized; a missing previous state is replaced by zeros. This cached target state can precede the bonus token when its state was not saved. The current target distribution is used only for supervision and evaluation. Input dimensions are 2,880/5,760 for GPT-OSS and 2,048/4,096 for DSpark. Heads are linear or of the form $V\,\mathrm{SiLU}(Ux)+c$, with bottleneck width 8 or 32 and one (temperature) or two (binary) outputs. A scalar output predicts $T=\exp(\mathrm{clamp}(z,-7,4))$ and minimizes $\ell_{\mathrm{TV}}$ with $K=7$. The binary head uses label $\mathbf{1}[A_i(1)>A_i(0)]$ and cross-entropy weighted by $0.8^{i-1}|A_i(1)-A_i(0)|$, normalized by the mean training weight, at draft position $i$. Its predicted probability for $T=1$ is thresholded over $\{0,0.025,\ldots,1\}\cup\{1.001\}$ on validation. Each head family and input variant selects its hyperparameters, positive-epoch checkpoint, and binary decision threshold on that task's validation $L_{\mathrm{pos}}$.

\paragraph{Optimization and selection.}
Both matrix and head training use Adam with learning rates $\{10^{-3},10^{-2},5\times10^{-2}\}$, $(\beta_1,\beta_2)=(0.9,0.999)$, $\epsilon=10^{-8}$, batch size 8,192, and 32 epochs, with no weight decay or learning-rate schedule. Candidate checkpoints are taken after 1, 2, 4, 8, 16, 24, and 32 epochs. Matrices start from identity or a diagonal calibration selected by task-averaged validation $L_{\mathrm{pos}}$; temperature heads start from $T=1$ or a task-specific validation-selected scalar.\footnote{The scalar initialization uses the block-wise replay score $1+\sum_{i=1}^k\prod_{j=1}^i A_j$, averaged over blocks with uncached positions set to zero, and is bounded below by $0.2$. Final head selection uses $L_{\mathrm{pos}}$.} Binary heads start from the utility-weighted training label frequency. Learned choices are frozen before test evaluation.

\paragraph{Oracle comparators.}
The scalar token-adaptive oracle selects one temperature per task from $\mathcal G=\{0,0.2,0.4,0.6,0.8,1,1.2,1.4,2\}$ by test $L_{\mathrm{pos}}$. \modified{The deployment baseline uses the same grid with validation selection, and diagonal scaling uses the six-point grid above.} Token oracles instead maximize $A(T)$ separately at each eligible cached position over $\mathcal G$ or $\{0,1\}$, with $A(0)=p(\arg\max_x q(x))$. They use the current target distribution and provide diagnostic upper bounds within their respective grids on these cached contexts; they are not deployable heads.

\subsection{\modified{Serving evaluation protocol}}\label{appendix:serving_protocol}
\modified{This section gives the online serving settings for Table~\ref{tab:serving_results} and Appendix~\ref{appendix:serving_results}. We compare speculative decoding at $T_d=1$ with draft temperature scaling. For each model pair, one shared temperature maximizes the equal-task validation replay $L_{\mathrm{pos}}$ over the scalar grid and is frozen before online test evaluation: 0.4, 0.4, and 0.6 for GPT-OSS/EAGLE-3, GPT-OSS/DFlash, and Qwen3.6/DSpark. This serving protocol uses one temperature across tasks, while the calibration-family comparisons in Tables~\ref{tab:dirichlet-scaling-results} and~\ref{tab:token-adaptive-temperature-scaling} select temperatures separately for each task.}

\modified{Each evaluation uses two NVIDIA GH200 GPUs, with one independent tensor-parallel-size-one server per GPU. Total client concurrency is drawn from $\{16,64,128\}$ across both replicas; ordered request IDs are routed alternately to them. Per-server active-request limits are 64/64/20, KV-token budgets are 131,072/131,072/65,536, and static-memory fractions are 0.85/0.85/0.94 for the three model pairs, respectively. These capacity limits are unchanged across the experiments.}

\modified{We use the same 2,082 held-out requests as the replay test partition (661 GSM8K, 250 MATH-500, 15 AIME25, 83 HumanEval, 130 MBPP, 528 LiveCodeBench, 40 MT-Bench, and 375 Arena-Hard v2), now with fresh generation rather than cached trajectories. GPT-OSS uses low reasoning, and Qwen3.6 disables thinking. Draft lengths remain 9/9/8; target temperature is 1, top-$p$ is 1, top-$k$ is unrestricted, and the output cap is 2,048 tokens. We use 3 seeds for each test. 
}

\modified{Server startup and warm-up are excluded. After warm-up, server caches are flushed; task timing includes fresh prefill, generation, and completion of all requests. Distribution instrumentation and trajectory caching are disabled in both conditions. For task $j$, let $N_j$, $V_j$, and $t_j$ denote output tokens, the sum of request-level verify counts, and measured elapsed seconds. Per-task accepted length is $L_j=N_j/V_j$ and output throughput is $Q_j=N_j/t_j$. We report the equal-task means $\bar L=\frac18\sum_j L_j$ and $\bar Q=\frac18\sum_j Q_j$. For either metric $M$, the gain within each seed is $100(M_{\mathrm{cal}}/M_{\mathrm{original}}-1)$. We report means and standard errors (SE, sample standard deviation divided by $\sqrt{3}$) of the absolute values and paired gains across the three seeds. Non-streaming request latency is measured from HTTP dispatch to response completion, including server queuing, prefill, and generation, and excluding time waiting for a client worker. For each seed, p50 and p95 pool all 2,082 request records per condition; the resulting quantiles and percentage reductions are then summarized across seeds.}
\section{Additional Experiment Results}

\subsection{Observations for Underconfident Draft Models}\label{appendix:observations}

\begin{figure}[t]
    \centering
    \includegraphics[width=\linewidth]{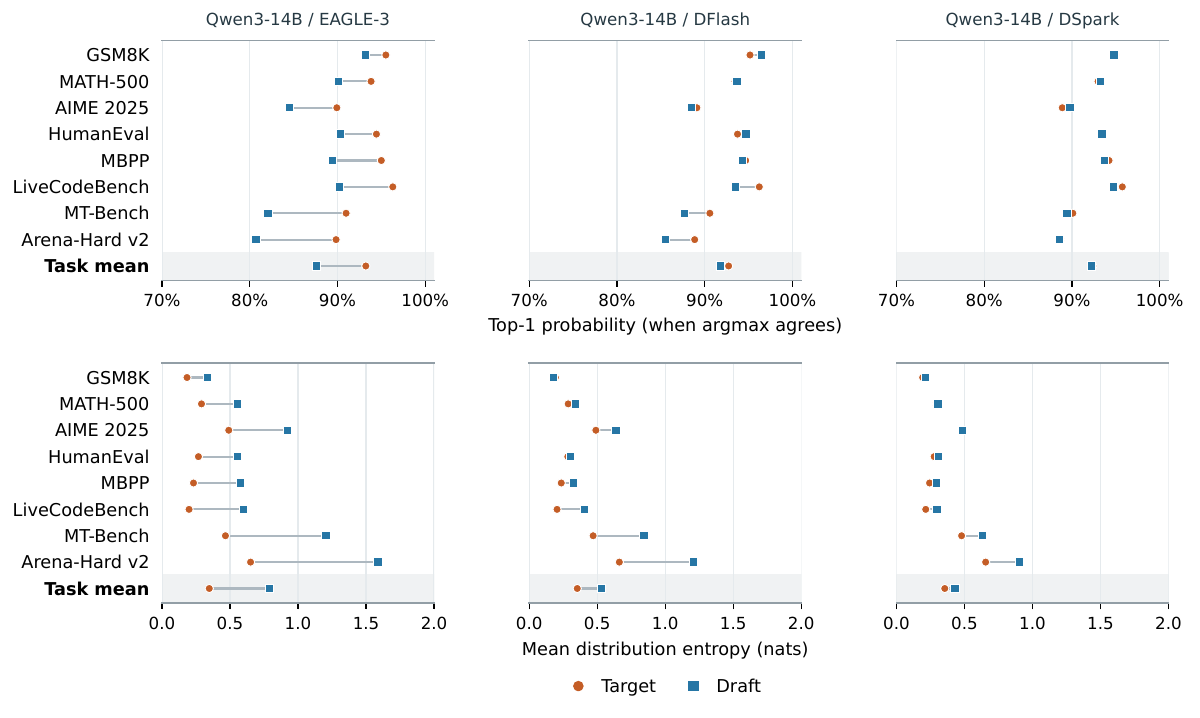}
    \vspace{-0.5cm}
    \caption{\modified{Draft--target distribution diagnostics for the DeepSeek-trained Qwen3-14B checkpoints, following Figure~\ref{fig:matched_top1_prob_ent}. All three have smaller task-mean probability and entropy gaps than the corresponding model pairs in Figure~\ref{fig:matched_top1_prob_ent}.}}
    \label{fig:deepseek_distribution_gap}
    \vspace{-0.1cm}
\end{figure}

\begin{figure}[H]
    \centering
    \includegraphics[width=\linewidth]{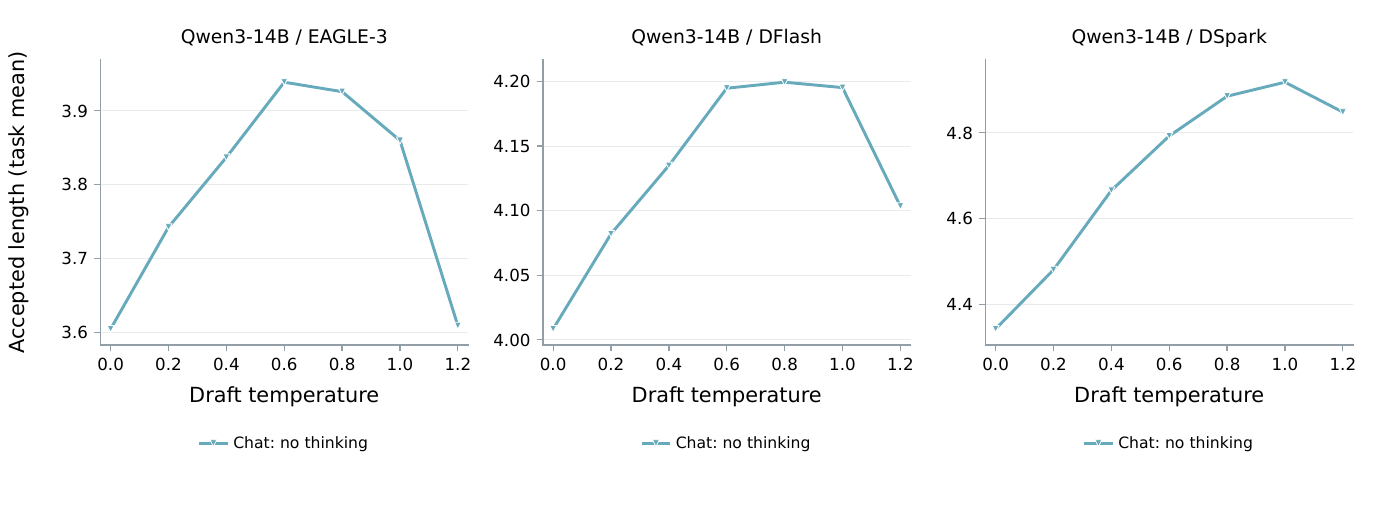}
    \vspace{-0.5cm}
    \caption{Temperature scaling for the three DeepSeek-trained Qwen3-14B draft models. Each curve is the accepted length averaged equally across the eight benchmarks, with thinking disabled and target temperature fixed at $T_t=1$.}
    \label{fig:temperature_scaling_deepseek_qwen3_14b}
    \vspace{-0.1cm}
\end{figure}

\paragraph{\modified{Distribution gaps for DeepSeek-trained draft models.}}
\modified{We apply the diagnostics in Figure~\ref{fig:matched_top1_prob_ent} to the same Qwen3-14B checkpoints as Figure~\ref{fig:temperature_scaling_deepseek_qwen3_14b}, using $T_d=T_t=1$, seven draft positions, thinking disabled, and the same 4,155 prompts with a 2,048-token cap. Figure~\ref{fig:deepseek_distribution_gap} uses cached SpecForge/SGLang trajectories: top-1 probabilities are conditioned on argmax agreement at eligible positions, and entropy uses all eligible positions. Both metrics are averaged within each task and then equally across the eight tasks, as in Figure~\ref{fig:matched_top1_prob_ent}.}

\modified{For EAGLE-3, DFlash, and DSpark, respectively, the matched-top-1 probability gaps are 5.65, 0.93, and 0.18 percentage points, compared with 11.94, 9.86, and 7.35 for the corresponding drafters in Figure~\ref{fig:matched_top1_prob_ent}. Their draft-minus-target entropy gaps are 0.443, 0.177, and 0.075 nats, compared with 0.842, 0.641, and 0.744; their mean full-distribution $\ell_1$ gaps are also smaller, at 0.470, 0.422, and 0.316 versus 0.664, 0.539, and 0.474. At a fixed context, the acceptance gain from calibration is exactly half its reduction in the $\ell_1$ gap (Eq.~\ref{eq:acceptance_probability}). The smaller probability-scale mismatch leaves less underconfidence for global sharpening to correct, consistent with Figure~\ref{fig:temperature_scaling_deepseek_qwen3_14b}; EAGLE-3 retains more underconfidence and shows the largest gain among the three.}

\paragraph{Temperature scaling for DeepSeek-trained draft models.}
We additionally evaluate DeepSeek's Qwen3-14B draft checkpoints for \href{https://huggingface.co/deepseek-ai/eagle3_qwen3_14b_ttt7}{EAGLE-3}, \href{https://huggingface.co/deepseek-ai/dflash_qwen3_14b_block7}{DFlash}, and \href{https://huggingface.co/deepseek-ai/dspark_qwen3_14b_block7}{DSpark}. These online sweeps use our modified DeepSpec evaluator with independent draft-temperature control, $T_t=1$, seven draft positions, thinking disabled, and a 2,048-token generation cap. DSpark's confidence head is disabled. We use all 4,155 prompts from the eight benchmarks in Table~\ref{tab:dataset_settings} and average the recorded accepted lengths equally across tasks.

Figure~\ref{fig:temperature_scaling_deepseek_qwen3_14b} shows limited gains from a global temperature correction for these checkpoints. EAGLE-3 improves from $3.860$ at $T_d=1$ to $3.939$ at $T_d=0.6$ ($+2.05\%$). DFlash is nearly flat around its optimum: $4.195$ at $T_d=1$ versus $4.200$ at $T_d=0.8$ ($+0.10\%$). DSpark attains its highest task mean, $4.918$, at $T_d=1$. These optima are selected retrospectively from the displayed grid. 
Thus, different from the draft models shown in Figure~\ref{fig:temperature_scaling_accepted_length}, these three draft models {are generally better calibrated}, and temperature scaling only gives marginal help. It will be an interesting future direction to find out what the secret sauce is inside this, which may help with further exploration of token-adaptive draft calibration.

\subsection{Offline Trajectory Replay}\label{appendix:offline_replay}
As plotted in Figure~\ref{fig:replay_illustration}, we here illustrate the trajectory replay process in words. To prepare a trajectory for replay, we first generate a sequence with the desired target and draft models with a reference calibration strategy, e.g., the standard temperature choice $T_d = T_t$. During the generation, we save {the final hidden states of draft and target models at eligible reference positions, where all preceding proposals were accepted}, which can be obtained during draft generation and the verification process.
Then, when doing replay, at each draft position, we input the final hidden states to the LM Head to recover draft and target logits. Post-hoc calibration strategies are then applied to draft logits to obtain the calibrated draft distribution $q$, and compute the acceptance probability based on (\ref{eq:acceptance_probability}) to test the calibration strategy. The pipeline is illustrated in Figure~\ref{fig:replay_illustration}.

\paragraph{Replay--online comparison.}
We compare full-task temperature replay against the historical online sweeps for the same three model pairs and low/non-thinking settings used in Appendix~\ref{appendix:replay_training}, with $T_t=1$ and $T_d\in\{0,0.2,0.4,0.6,0.8,1,1.2,1.4\}$. Each comparison uses all 4,155 prompts across the eight tasks, rather than the test-only partition used in the calibration tables. Replay evaluates the fixed $T_d=1$ reference contexts; online runs generate fresh proposals at each temperature. We use runs without distribution instrumentation where available; the $T_d=1$ online reference is instrumented.

\begin{figure}[t]
    \centering
    \includegraphics[width=\linewidth]{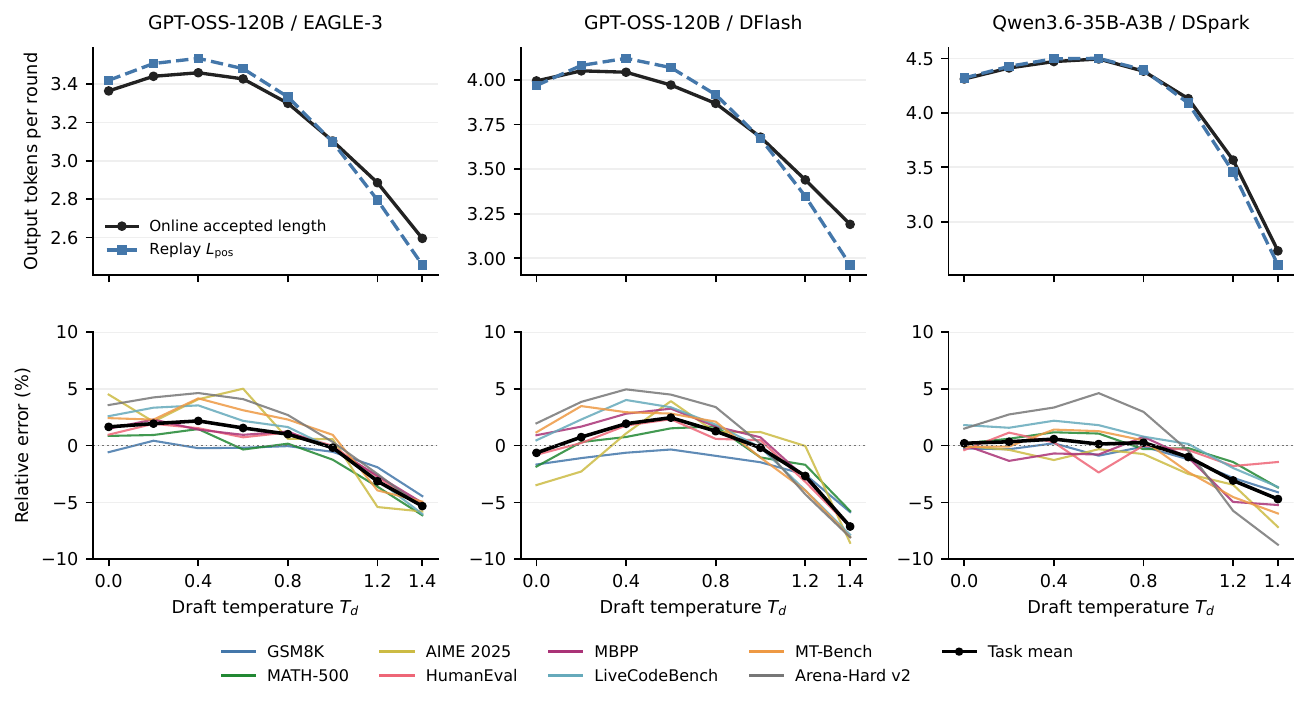}
    \caption{Temperature replay versus online decoding. Top: online accepted length and replay $L_{\mathrm{pos}}$, each averaged equally across tasks. Bottom: signed relative error $100(L_{\mathrm{pos}}-L_{\mathrm{online}})/L_{\mathrm{online}}$, with colored curves for individual tasks and black curves computed from the task means. Target temperature is fixed at $T_t=1$; draft lengths are $K=9,9,8$ from left to right. These are full-task diagnostics, separate from the held-out calibration tables.}
    \label{fig:replay_online_gap}
\end{figure}

Across the 64 task--temperature combinations per model pair, mean absolute relative errors are 2.42\%, 2.64\%, and 1.83\% for EAGLE-3, DFlash, and DSpark, respectively; maximum errors are 6.13\%, 8.56\%, and 8.72\%. The task-mean optimal temperatures agree for EAGLE-3 ($0.4$) and DSpark ($0.6$); for DFlash, replay selects $0.4$ versus the online optimum $0.2$, costing only $0.008$ output tokens per verification round on the online task mean. Thus replay captures the broad benefit of sharpening, but fixed reference contexts, reference-policy eligibility, and products of mean position-wise acceptance rates can bias its estimates. We use it as a diagnostic proxy rather than an exact prediction of online acceptance or throughput.

\subsection{\modified{Acceptance gains, throughput, and latency}}\label{appendix:serving_results}
\modified{These results use the same settings and held-out test split across all eight tasks as Table~\ref{tab:serving_results}, with the full protocol in Appendix~\ref{appendix:serving_protocol}. Table~\ref{tab:serving_results} places accepted length and throughput before and after calibration side by side. Table~\ref{tab:serving_uncertainty} gives the corresponding mean $\pm$ SE values. Both improve on average in all nine conditions: accepted length by 8.82\%--10.89\% and throughput by 5.13\%--22.08\%. The absolute difference between their mean gains has a median of 0.72 percentage points and a maximum of 13.26 points, with the largest difference at DSpark concurrency 64.}

\modified{For each task $j$, the identity $Q_j=L_j(V_j/t_j)$ shows that throughput and accepted length improve proportionally when the effective verification rate $V_j/t_j$ stays fixed. Continuous batching changes the number and context lengths of requests processed together as requests enter and finish, so different sampled outputs can change this rate even with identical server settings. Prefill, scheduling, and completion of the final requests also contribute to elapsed time, allowing throughput gains to differ despite negligible temperature-scaling overhead. Equal-task averaging aligns the reporting convention, although ratios of means need not preserve identical task-level gain relationships.}

\modified{Table~\ref{tab:serving_latency} gives request-latency quantiles, and Table~\ref{tab:serving_task_results} reports all 72 task-level comparisons, each as mean $\pm$ SE across seeds. Mean throughput improves in all nine aggregate conditions and 70 of the 72 task comparisons. Across individual seeds, 26 of 27 aggregate comparisons improve; the remaining change is $-0.56\%$ for EAGLE-3 at concurrency 128. Mean p50 and p95 latency reductions are positive in all nine conditions.}

\begin{table}[t]
\centering
\footnotesize
\setlength{\tabcolsep}{1.4pt}
\caption{\modified{Non-streaming request latency in seconds, mean $\pm$ SE over the same three seeds as Table~\ref{tab:serving_results}. Parentheses show mean $\pm$ SE of per-seed percentage reductions. Quantiles pool 2,082 requests within each seed; timing includes server queuing, prefill, and generation after dispatch, excluding time waiting for a client worker.}}
\label{tab:serving_latency}
\modified{
\resizebox{\linewidth}{!}{\begin{tabular}{@{}lrrrrr@{}}
\toprule
& & \multicolumn{2}{c}{p50 latency (s)} & \multicolumn{2}{c}{p95 latency (s)} \\
\cmidrule(lr){3-4}\cmidrule(l){5-6}
Target / drafter & $C$ & Original & Calibrated & Original & Calibrated \\
\midrule
\multirow{3}{*}{\makecell[l]{GPT-OSS-120B\\EAGLE-3}} & 16 & $7.68\!\pm\!0.21$ & $6.97\!\pm\!0.02\;(9.16\!\pm\!2.26)\%$ & $56.02\!\pm\!0.65$ & $51.00\!\pm\!2.09\;(8.99\!\pm\!3.10)\%$ \\
 & 64 & $11.44\!\pm\!0.24$ & $10.01\!\pm\!0.10\;(12.43\!\pm\!1.19)\%$ & $76.71\!\pm\!0.93$ & $69.32\!\pm\!0.24\;(9.61\!\pm\!0.77)\%$ \\
 & 128 & $13.60\!\pm\!0.31$ & $12.45\!\pm\!0.11\;(8.29\!\pm\!2.50)\%$ & $87.33\!\pm\!0.81$ & $82.22\!\pm\!0.72\;(5.84\!\pm\!0.21)\%$ \\
\midrule
\multirow{3}{*}{\makecell[l]{GPT-OSS-120B\\DFlash}} & 16 & $4.10\!\pm\!0.09$ & $3.87\!\pm\!0.07\;(5.60\!\pm\!1.95)\%$ & $26.60\!\pm\!0.33$ & $24.19\!\pm\!0.23\;(9.00\!\pm\!1.40)\%$ \\
 & 64 & $7.17\!\pm\!0.06$ & $6.65\!\pm\!0.10\;(7.30\!\pm\!1.26)\%$ & $44.41\!\pm\!0.17$ & $39.50\!\pm\!0.60\;(11.05\!\pm\!1.38)\%$ \\
 & 128 & $9.51\!\pm\!0.06$ & $8.69\!\pm\!0.09\;(8.64\!\pm\!1.41)\%$ & $56.40\!\pm\!0.41$ & $50.87\!\pm\!0.23\;(9.81\!\pm\!0.82)\%$ \\
\midrule
\multirow{3}{*}{\makecell[l]{Qwen3.6-35B-A3B\\DSpark}} & 16 & $5.35\!\pm\!0.04$ & $4.99\!\pm\!0.14\;(6.69\!\pm\!3.37)\%$ & $34.03\!\pm\!0.56$ & $31.40\!\pm\!0.51\;(7.66\!\pm\!2.26)\%$ \\
 & 64 & $12.43\!\pm\!0.33$ & $11.01\!\pm\!0.31\;(11.26\!\pm\!3.80)\%$ & $63.78\!\pm\!3.27$ & $53.21\!\pm\!2.31\;(16.11\!\pm\!5.69)\%$ \\
 & 128 & $21.53\!\pm\!0.13$ & $18.93\!\pm\!0.93\;(12.09\!\pm\!4.04)\%$ & $115.99\!\pm\!9.68$ & $97.07\!\pm\!0.87\;(15.11\!\pm\!7.18)\%$ \\
\bottomrule
\end{tabular}}}
\end{table}

\begin{table}[htbp]
\centering
\footnotesize
\setlength{\tabcolsep}{3pt}
\caption{\modified{Per-task online accepted length (AL) and output-throughput gains (\%) over $T_d=1$, mean $\pm$ SE across three seeds. Each seed contributes a calibrated-to-original ratio minus one. AIME25 has only 15 requests and MT-Bench 40, so their realized in-flight loads cannot reach every configured concurrency.}}
\label{tab:serving_task_results}
\modified{
\begin{tabular}{@{}lrrrrrr@{}}
\toprule
& \multicolumn{2}{c}{$C=16$} & \multicolumn{2}{c}{$C=64$} & \multicolumn{2}{c}{$C=128$} \\
Task & AL & Throughput & AL & Throughput & AL & Throughput \\
\midrule
\multicolumn{7}{c}{GPT-OSS-120B / EAGLE-3} \\
\midrule
GSM8K & $7.95\!\pm\!0.24$ & $11.42\!\pm\!3.51$ & $7.59\!\pm\!0.57$ & $9.14\!\pm\!4.67$ & $7.31\!\pm\!0.99$ & $2.04\!\pm\!11.08$ \\
MATH-500 & $11.32\!\pm\!0.64$ & $13.86\!\pm\!2.59$ & $11.52\!\pm\!0.58$ & $13.97\!\pm\!4.69$ & $12.17\!\pm\!0.26$ & $11.61\!\pm\!4.05$ \\
AIME25 & $10.73\!\pm\!2.02$ & $-2.10\!\pm\!3.63$ & $10.73\!\pm\!2.02$ & $7.92\!\pm\!9.91$ & $11.01\!\pm\!1.77$ & $-4.71\!\pm\!4.30$ \\
HumanEval & $12.53\!\pm\!0.74$ & $7.54\!\pm\!5.28$ & $12.40\!\pm\!0.69$ & $9.88\!\pm\!5.69$ & $12.25\!\pm\!0.46$ & $4.33\!\pm\!3.11$ \\
MBPP & $12.14\!\pm\!0.31$ & $9.62\!\pm\!1.23$ & $11.80\!\pm\!0.46$ & $14.75\!\pm\!1.26$ & $12.46\!\pm\!0.17$ & $6.78\!\pm\!2.75$ \\
LiveCodeBench & $12.70\!\pm\!0.62$ & $10.47\!\pm\!2.66$ & $13.23\!\pm\!0.54$ & $13.82\!\pm\!2.53$ & $12.94\!\pm\!0.28$ & $5.54\!\pm\!0.34$ \\
MT-Bench & $9.59\!\pm\!0.37$ & $18.09\!\pm\!5.37$ & $10.88\!\pm\!1.02$ & $6.27\!\pm\!3.90$ & $11.35\!\pm\!0.41$ & $5.72\!\pm\!5.92$ \\
Arena-Hard v2 & $8.58\!\pm\!0.46$ & $10.18\!\pm\!5.30$ & $8.58\!\pm\!0.59$ & $8.04\!\pm\!0.70$ & $8.85\!\pm\!0.40$ & $5.27\!\pm\!1.50$ \\
\midrule
\multicolumn{7}{c}{GPT-OSS-120B / DFlash} \\
\midrule
GSM8K & $6.93\!\pm\!0.51$ & $9.64\!\pm\!4.04$ & $6.33\!\pm\!0.44$ & $9.33\!\pm\!1.96$ & $6.39\!\pm\!0.88$ & $4.21\!\pm\!9.46$ \\
MATH-500 & $8.89\!\pm\!0.41$ & $7.09\!\pm\!2.24$ & $9.15\!\pm\!0.47$ & $9.84\!\pm\!1.36$ & $9.46\!\pm\!0.85$ & $9.63\!\pm\!2.85$ \\
AIME25 & $9.62\!\pm\!3.12$ & $13.80\!\pm\!4.80$ & $9.62\!\pm\!3.12$ & $13.32\!\pm\!6.93$ & $8.61\!\pm\!3.55$ & $12.86\!\pm\!8.92$ \\
HumanEval & $10.08\!\pm\!0.33$ & $10.51\!\pm\!1.35$ & $9.05\!\pm\!0.38$ & $5.74\!\pm\!4.08$ & $9.86\!\pm\!0.21$ & $8.33\!\pm\!2.26$ \\
MBPP & $10.54\!\pm\!0.40$ & $9.48\!\pm\!3.61$ & $10.95\!\pm\!0.54$ & $9.50\!\pm\!3.02$ & $10.58\!\pm\!0.52$ & $10.66\!\pm\!1.34$ \\
LiveCodeBench & $10.97\!\pm\!0.49$ & $8.80\!\pm\!2.27$ & $11.33\!\pm\!0.75$ & $12.28\!\pm\!1.04$ & $11.53\!\pm\!0.61$ & $12.43\!\pm\!0.67$ \\
MT-Bench & $8.29\!\pm\!0.82$ & $13.36\!\pm\!3.31$ & $9.04\!\pm\!0.39$ & $11.36\!\pm\!5.50$ & $9.17\!\pm\!1.25$ & $12.52\!\pm\!4.82$ \\
Arena-Hard v2 & $9.55\!\pm\!0.34$ & $9.70\!\pm\!2.20$ & $9.75\!\pm\!0.47$ & $10.88\!\pm\!0.92$ & $9.49\!\pm\!0.58$ & $9.08\!\pm\!0.92$ \\
\midrule
\multicolumn{7}{c}{Qwen3.6-35B-A3B / DSpark} \\
\midrule
GSM8K & $8.18\!\pm\!0.84$ & $11.42\!\pm\!5.18$ & $8.62\!\pm\!0.99$ & $26.72\!\pm\!5.06$ & $8.45\!\pm\!0.95$ & $18.60\!\pm\!3.53$ \\
MATH-500 & $7.17\!\pm\!0.10$ & $9.93\!\pm\!3.12$ & $6.75\!\pm\!0.29$ & $16.90\!\pm\!8.72$ & $6.74\!\pm\!0.27$ & $8.77\!\pm\!2.55$ \\
AIME25 & $10.52\!\pm\!0.93$ & $12.26\!\pm\!4.73$ & $10.41\!\pm\!1.03$ & $14.31\!\pm\!6.28$ & $10.63\!\pm\!1.14$ & $22.59\!\pm\!14.03$ \\
HumanEval & $9.61\!\pm\!2.66$ & $25.04\!\pm\!4.05$ & $7.77\!\pm\!0.18$ & $46.00\!\pm\!11.43$ & $7.36\!\pm\!0.39$ & $29.93\!\pm\!4.90$ \\
MBPP & $12.26\!\pm\!0.61$ & $13.51\!\pm\!7.91$ & $10.47\!\pm\!2.74$ & $34.93\!\pm\!10.08$ & $11.22\!\pm\!1.71$ & $57.42\!\pm\!14.49$ \\
LiveCodeBench & $7.68\!\pm\!0.14$ & $7.00\!\pm\!2.02$ & $7.43\!\pm\!0.11$ & $15.53\!\pm\!7.27$ & $7.60\!\pm\!0.13$ & $12.73\!\pm\!6.29$ \\
MT-Bench & $13.44\!\pm\!1.09$ & $9.76\!\pm\!5.09$ & $11.95\!\pm\!0.27$ & $11.58\!\pm\!4.49$ & $11.70\!\pm\!0.44$ & $22.19\!\pm\!17.10$ \\
Arena-Hard v2 & $8.44\!\pm\!0.72$ & $10.89\!\pm\!1.96$ & $8.45\!\pm\!0.30$ & $15.05\!\pm\!3.01$ & $8.41\!\pm\!0.34$ & $13.82\!\pm\!5.51$ \\
\bottomrule
\end{tabular}}
\end{table}

\begin{table}[htbp]
\centering
\small
\setlength{\tabcolsep}{4pt}
\caption{\modified{Complete serving results for Table~\ref{tab:serving_results}, mean $\pm$ standard error (SE) over three sampling seeds. AL denotes accepted length. Parentheses show percentage gains, computed within each seed before averaging; their SEs are in percentage points. Both metrics average the eight tasks equally within each seed. $C$ is total concurrency across two single-GPU replicas.}}
\label{tab:serving_uncertainty}
\modified{
\begin{tabular}{@{}lrrrrr@{}}
\toprule
& & \multicolumn{2}{c}{AL} & \multicolumn{2}{c}{Throughput (output tok/s)} \\
\cmidrule(lr){3-4}\cmidrule(l){5-6}
Target / drafter & $C$ & Original & Calibrated (gain) & Original & Calibrated (gain) \\
\midrule
\multirow{6}{*}{\makecell[l]{GPT-OSS-120B\\EAGLE-3}} & 16 & $3.102\pm0.004$ & \makecell[r]{$3.432\pm0.003$\\$(10.64\pm0.19)\%$} & $760.9\pm11.6$ & \makecell[r]{$836.2\pm8.9$\\$(9.92\pm1.02)\%$} \\
 & 64 & $3.104\pm0.005$ & \makecell[r]{$3.436\pm0.008$\\$(10.71\pm0.28)\%$} & $1582.6\pm25.9$ & \makecell[r]{$1756.3\pm30.3$\\$(11.01\pm2.03)\%$} \\
 & 128 & $3.103\pm0.003$ & \makecell[r]{$3.441\pm0.009$\\$(10.89\pm0.36)\%$} & $2141.6\pm32.5$ & \makecell[r]{$2249.7\pm38.6$\\$(5.13\pm3.02)\%$} \\
\midrule
\multirow{6}{*}{\makecell[l]{GPT-OSS-120B\\DFlash}} & 16 & $3.657\pm0.009$ & \makecell[r]{$3.996\pm0.011$\\$(9.27\pm0.45)\%$} & $1449.5\pm8.3$ & \makecell[r]{$1592.5\pm6.0$\\$(9.87\pm0.29)\%$} \\
 & 64 & $3.656\pm0.011$ & \makecell[r]{$3.994\pm0.004$\\$(9.25\pm0.44)\%$} & $2765.9\pm26.5$ & \makecell[r]{$3034.3\pm61.8$\\$(9.68\pm1.29)\%$} \\
 & 128 & $3.660\pm0.011$ & \makecell[r]{$3.998\pm0.015$\\$(9.26\pm0.64)\%$} & $3548.1\pm22.5$ & \makecell[r]{$3866.0\pm66.8$\\$(8.98\pm2.32)\%$} \\
\midrule
\multirow{6}{*}{\makecell[l]{Qwen3.6-35B-A3B\\DSpark}} & 16 & $4.049\pm0.010$ & \makecell[r]{$4.434\pm0.006$\\$(9.51\pm0.32)\%$} & $1174.9\pm37.5$ & \makecell[r]{$1313.9\pm22.2$\\$(12.04\pm3.86)\%$} \\
 & 64 & $4.066\pm0.022$ & \makecell[r]{$4.425\pm0.009$\\$(8.82\pm0.39)\%$} & $1858.6\pm37.3$ & \makecell[r]{$2265.8\pm74.1$\\$(22.08\pm5.64)\%$} \\
 & 128 & $4.070\pm0.021$ & \makecell[r]{$4.430\pm0.012$\\$(8.84\pm0.29)\%$} & $1782.9\pm33.2$ & \makecell[r]{$2130.7\pm33.7$\\$(19.53\pm1.15)\%$} \\
\bottomrule
\end{tabular}}
\end{table}

\end{document}